\documentclass[a4paper,fleqn]{cas-sc}

\usepackage[authoryear,longnamesfirst]{natbib}

\def\tsc#1{\csdef{#1}{\textsc{\lowercase{#1}}\xspace}}
\tsc{WGM}
\tsc{QE}

\begin{document}
\let\WriteBookmarks\relax
\def\floatpagepagefraction{1}
\def\textpagefraction{.001}

\shorttitle{Brain-Aligned Multi-Stream Video Transformers with Sparse Self-Selection}    

\shortauthors{A.H. Fadaei et~al.}  

\title [mode = title]{Brain-Aligned Multi-Stream Video Transformers with Sparse Self-Selection}  



%

\author[1]{Amir Hosein Fadaei}[orcid=0000-0002-5989-1093]

\ead{hosein.fadaei@ut.ac.ir}


\credit{Conceptualization, Data curation, Methodology, Software, Investigation, Formal analysis, Validation, Visualization, Writing - original draft, Writing - review and editing}

\affiliation[1]{organization={University of Tehran, Faculty of Electrical and Computer Engineering},
            addressline={North Kargar st.}, 
            city={Tehran},
            postcode={1439957131}, 
            state={Tehran},
            country={Iran}}

\author[1]{Mahyar Maleki}[orcid=0009-0002-0411-2256]

\ead{mahyar.maleki@ut.ac.ir}

\credit{Data curation, Investigation, Formal analysis}

\author[1]{Mohammad-Reza A. Dehaqani}[orcid=0000-0003-4365-4365]

\ead{dehaqani@ut.ac.ir}

\cormark[1]

\credit{Conceptualization, Funding acquisition, Project administration, Supervision, Validation, Writing - review and editing}

\cortext[1]{Corresponding author}


\begin{abstract}
Modern video transformers typically ignore principles from primate vision and are rarely evaluated against neural data, limiting their biological interpretability. We introduce a sparse winner-takes-all token selection module that replaces dense self-attention to improve efficiency and approximate competitive routing observed in biological visual circuits. We further propose a neuro-inspired split-and-fuse video transformer which uses two complementary pathways: a high-resolution, low-frame-rate "what" stream and a low-resolution, high-frame-rate "where" stream, fused before classification. On Kinetics-400 and Something-Something V2, our best variant operates on the Pareto frontier of accuracy versus inference time among models of comparable scale and pretraining, and showing improved robustness to spatial perturbations. Using representational similarity analysis between model embeddings and time-resolved EEG recordings for the same video stimuli, our model attains a peak brain-model correlation of 0.18 (about 78\% of the noise ceiling) and consistently outperforms strong video transformer baselines, suggesting that pathway specialization and sparse competition are useful inductive biases for efficient, brain-aligned video understanding.
\end{abstract}


\begin{highlights}
\item Introduces a neuro-inspired split-and-fuse video transformer with separate "what" and "where" streams whose complementary EEG alignment patterns are consistent with ventral and dorsal visual pathway specializations.	
\item Replaces dense self-attention with a sparse winner-takes-all selection module, inspired by competitive routing observed in biological visual circuits, for efficient video processing.
\item Achieves competitive or state-of-the-art accuracy and pareto frontier accuracy-latency trade-offs on Kinetics-400 and Something-Something V2 among models of comparable scale and training.
\item Demonstrates improved robustness to spatial perturbations compared to strong video transformer baselines.
\item Shows higher EEG representational similarity than existing video models, indicating more brain-aligned visual representations.
\end{highlights}


\maketitle

\section{Introduction}

Modern video understanding models have achieved impressive performance on large-scale benchmarks, yet they are typically developed with little regard for how biological visual systems process dynamic scenes~\citep{yamins2016, celeghin2023}. Existing transformer-based architectures for video recognition often rely on a single stream with dense self-attention over all space-time tokens, which offers strong expressivity but leads to quadratic scaling in sequence length and limited interpretability in relation to primate vision~\citep{Bertasius2021,Arnab2021,Liu2022}. At the same time, a growing NeuroAI literature suggests that models whose internal representations align better with neural data can exhibit improved robustness and may provide more mechanistic insights into perception, but systematic evaluations of video transformers against human brain activity remain scarce~\citep{schrimpf2018,dapello2022,conwell2024large}.

In contrast, primate vision relies on multiple specialized pathways that process complementary aspects of the visual input. The ventral "what" pathway is associated with detailed object recognition, whereas the dorsal "where" pathway emphasizes motion, spatial relations, and action-related information ~\citep{goodale1992separate,creem2001}. This division of labor suggests that explicit pathway specialization could serve as a useful inductive bias for artificial systems that must understand complex video streams~\citep{Simonyan2014,Zareh2024}. However, most current video transformers process inputs in a single pathway and do not exploit a structured separation between high-resolution spatial processing and high-temporal-resolution motion processing~\citep{Bertasius2021,Liu2022}.

Another gap concerns computational efficiency and biological plausibility in attention mechanisms. Standard self-attention forms a dense weighted sum over all tokens for each query, which is computationally expensive and difficult to reconcile with evidence for sparse, competitive interactions in neural circuits~\citep{itti2001computational}. Neurophysiology and computational models point to winner-takes-all dynamics and lateral inhibition, in which strongly activated units suppress their neighbors so that information is routed through a relatively small subset of "winning" units~\citep{desimone1995neural,zhou2000coding}. Existing efficient attention variants partially address computational cost through architectural tricks or token pruning, but they rarely tie sparsity to a competitive selection mechanism and are even less often evaluated in terms of brain-model alignment~\citep{Rao2021,beltagy2020,kitaev2020}.

In this work, we introduce a neuro-inspired split-and-fuse video transformer that explicitly decomposes video processing into two parallel streams and incorporates sparse competitive token selection. The first stream ("what") operates at high spatial resolution and lower frame rate, prioritizing fine-grained object and texture information. The second stream ("where") operates at lower spatial resolution and higher frame rate, emphasizing motion and spatial dynamics. The two streams are processed by separate transformer encoders and fused before classification, either through simple feature concatenation or lightweight cross-attention. This design is motivated by the ventral/dorsal distinction in biological vision and is intended to test whether such pathway specialization can improve both engineering performance and alignment with human neural responses.

To further reduce computational cost and introduce a biologically motivated inductive bias, we replace dense self-attention with a sparse winner-takes-all selection module in each transformer block. Instead of forming a full attention map and computing a weighted sum over all value tokens, the module selects a single most informative token per query position and forwards only that token’s representation. During training, this discrete routing is made learnable via a Gumbel-softmax-based~\citep{jang2017categorical} approximation, while at inference time it reduces to a simple argmax-and-gather operation. This mechanism enforces sparse competition in token space, substantially lowering component memory and compute while preserving representational capacity, and offers a conceptual link to competitive routing in biological circuits~\citep{Rao2021,desimone1995neural}.

We evaluate the proposed architecture along three axes. First, we assess video recognition performance and efficiency on Kinetics-400 and Something-Something V2, comparing our split-and-fuse models, with and without sparse selection, against strong convolutional and transformer-based baselines of comparable scale and pretraining~\citep{Carreira2017,goyal2017something,Bertasius2021}. Second, we study robustness to input variation by applying spatial perturbations such as color jitter, geometric transformations, and brightness changes at test time and measuring the resulting accuracy degradation~\citep{fadaei2024}. Third, to quantify brain-model alignment, we perform representational similarity analysis between internal model embeddings and time-resolved EEG recordings collected while human participants view naturalistic video clips~\citep{kriegeskorte2008,schirrmeister2017}. This allows us to ask not only whether the model aligns with aggregate neural responses, but also how its different streams and layers relate to activity over distinct scalp regions~\citep{Cichy2014,schirrmeister2017}.

Our results show that the split-and-fuse architecture with sparse selection achieves competitive or state-of-the-art accuracy among models of comparable scale and pretraining while operating on the Pareto frontier of accuracy versus inference time, and that multi-stream variants are more resilient to spatial perturbations than a single-stream TimeSformer baseline~\citep{Bertasius2021}. Moreover, the EEG-based analyses reveal modest but consistent improvements in brain-model correspondence, with the full model and its sparse variants exhibiting higher representational similarity to human EEG than monolithic video transformers. The spatial organization of these effects further suggests complementary roles for the "what" and "where" streams, in a manner broadly consistent with ventral and dorsal pathway specializations~\citep{goodale1992separate,Zareh2024}. Together, these findings support the view that pathway specialization and sparse competition are promising inductive biases for efficient video transformers that better align with human visual processing.

In summary, this work makes three contributions: (i) a neuro-inspired split-and-fuse video transformer with functionally specialized what/where streams, (ii) a sparse winner-takes-all token selection module that serves as an efficient alternative to dense self-attention, and (iii) a systematic EEG-based RSA evaluation of video transformer variants, including stream-wise and regional analyses of brain-model alignment.

\section{Related work}

\subsection{Video action recognition and video transformers}

Early deep models for video understanding were predominantly convolutional, extending 2D CNNs to the temporal domain via 3D convolutions or two-stream designs that separate RGB and optical flow inputs~\citep{Simonyan2014}. Examples include I3D~\citep{Carreira2017}, two-stream CNNs	\citep{Simonyan2014}, and later channel-separated convolutional networks, which demonstrated strong performance on datasets such as Kinetics-400~\citep{Carreira2017} and Something-Something V2~\citep{goyal2017something} but relied on hand-crafted motion inputs or computationally expensive spatiotemporal convolutions. More recent architectures such as SlowFast explicitly introduce dual pathways operating at different frame rates to capture slow semantic content and fast motion, establishing the value of multi-path temporal processing within a CNN framework~\citep{Feichtenhofer2019}.

Transformer-based models brought a different inductive bias to video recognition by treating frames as sequences of tokens and applying self-attention across space and time~\citep{Arnab2021}. TimeSFormer, ViViT, Video Swin, MViT, MTV, and VideoMAE extend image transformers to video using various factorized or hierarchical attention patterns, achieving strong recognition accuracy on large-scale benchmarks~\citep{Bertasius2021,Arnab2021,Liu2022,Li2022,yan2022,wang2023}. MTV and related multiview or multiscale transformers process videos through several views at different spatial and temporal scales, but each encoder still uses dense self-attention, so computational cost scales quadratically with the number of tokens~\citep{yan2022,fan2021Multiscale}.

More recent video transformers have scaled dramatically through massive web-scale pretraining on billions of unlabeled clips (or millions of hours), enabling foundational models with strong zero-shot transfer to diverse tasks. Models like VideoPrism, InternVideo2, Video Swin Transformer V2, UniFormerV2, and V-JEPA 2 leverage self-supervised learning on datasets exceeding 36M videos or 1M+ hours, achieving state-of-the-art results on Kinetics-400 and Something-Something V2 while supporting downstream applications from action recognition to robotics planning and scientific video analysis~\citep{wang2024internvideo2, zhao2024videoprism, liu2022swin, lu2024enhancing, assran2025v}. These advances emphasize joint spatiotemporal encoders pretrained via masked modeling, contrastive objectives, or predictive architectures, often with giant ViT backbones (e.g., ViT-g)~\citep{dosovitskiy2020image}, but still grapple with quadratic attention costs in long-sequence regimes-paving the way for sparsity and pathway innovations like ours.

\subsection{Sparse attention mechanisms}

Dense self-attention is flexible but expensive, and many variants have been proposed to improve its efficiency~\citep{beltagy2020,kitaev2020, choromanski2021rethinking}. Sparse Transformer~\citep{child2019generating}, Longformer~\citep{beltagy2020}, Reformer ~\citep{kitaev2020}, BigBird~\citep{zaheer2020big} and related architectures restrict attention patterns using local windows, hashing, or low-rank approximations~\citep{wang2020linformer}; other methods perform dynamic token sparsification, patch shifting, or windowed attention to reduce the number of active tokens, particularly in vision transformers~\citep{Rao2021, Liu2021Swin, ryoo2021tokenlearner}. In video models, temporal shift modules, token shift transformers, and grouped spatiotemporal shift operations approximate temporal interactions with minimal additional parameters, enabling efficient action recognition~\citep{lin2019tsm,fan2020rubiksnet,zhang2021token,li2023simple}. Mixture-of-experts layers represent another form of learned conditional computation, where different tokens or regions are routed to different experts with sparse gates~\citep{shazeer2017}.

These approaches show that sparsity and conditional computation can significantly reduce cost, but most still compute a (possibly structured) weighted sum over many tokens and are not directly motivated by competitive interactions in neural circuits~\citep{Rao2021,jang2017categorical}. In contrast, our self-selection block replaces dense self-attention with a winner-takes-all mechanism that routes each query to a single selected value token, implementing sparse competitive routing rather than dense averaging~\citep{desimone1995neural,jang2017categorical}. This yields a much simpler attention map, substantially lowering memory and compute compared to dense attention, while providing a direct abstraction of competitive selection and lateral inhibition phenomena in visual cortex~\citep{desimone1995neural,zhou2000coding}.

\subsection{Brain-model alignment}

There is longstanding interest in using deep networks as models of the visual cortex and in designing architectures that better mirror biological organization~\citep{yamins2016}. Work on object recognition has shown that goal-driven deep convolutional networks trained on large-scale image datasets can predict neural responses in primate inferior temporal cortex and human ventral visual areas to a notable degree~\citep{yamins2016,dapello2022}. More recently, studies have systematically examined which inductive biases (e.g., texture vs.~shape bias, global vs.~local processing, architectural choices) influence brain-model alignment, and have highlighted the importance of the similarity metric itself for assessing alignment~\citep{conwell2024large,soni2024conclusions,schrimpf2018}. Additional research has explored brain-score-style benchmarks, convolutional networks for vision neuroscience, and the use of compact deep networks like EEGNet~\citep{lawhern2018eegnet,schirrmeister2017} for EEG decoding and visualization~\citep{schrimpf2018, celeghin2023, schirrmeister2017}.

Representational similarity analysis (RSA) has become a standard tool for comparing neural and model representations across time and space~\citep{kriegeskorte2008}. Prior work has used RSA to relate model activations to fMRI, MEG, and scalp EEG recordings, revealing time-resolved correspondences in object recognition and other tasks~\citep{Cichy2014, cichy2019deep, kriegeskorte2008, schirrmeister2017}. For dynamic stimuli, recurrent architectures have been shown to better capture visual cortex dynamics than feedforward models~\citep{kietzmann2019recurrence}, motivating explicit temporal modeling in brain-aligned video systems. However, most studies focus on static images or object recognition, and relatively few examine video transformers or multi-stream architectures in relation to human EEG, especially with explicit modeling of ventral/dorsal-like streams~\citep{goodale1992separate,Zareh2024}. Moreover, existing efficient attention or token-sparsification methods are rarely evaluated for their impact on brain-model alignment~\citep{Rao2021}.

\subsection{Positioning of the present work}

Our contribution sits at the intersection of these lines of research. Compared to prior video transformers and multi-stream architectures~\citep{Bertasius2021,Feichtenhofer2019,Zareh2024}, we combine an explicitly ventral/dorsal-inspired what/where split with a sparse winner-takes-all token selection mechanism, and we evaluate this design on large-scale action recognition benchmarks with matched capacity baselines~\citep{Carreira2017,goyal2017something}. Compared to earlier NeuroAI and RSA-based alignment work~\citep{yamins2016,kriegeskorte2008,schrimpf2018, schirrmeister2017,Cichy2014}, we provide a systematic EEG evaluation of video transformer variants, including stream-wise and region-wise analyses that probe how pathway specialization and sparse competition affect both overall brain-model similarity and its spatial organization across scalp regions. This allows us to test whether neuro-inspired multi-stream design and competitive token routing serve as useful inductive biases not only for efficiency and robustness, but also for producing representations that are more aligned with human visual dynamics.

\section{Methodology}

\subsection{Model Architecture}

\begin{figure}
  \centering
    \includegraphics[width=.9\textwidth]{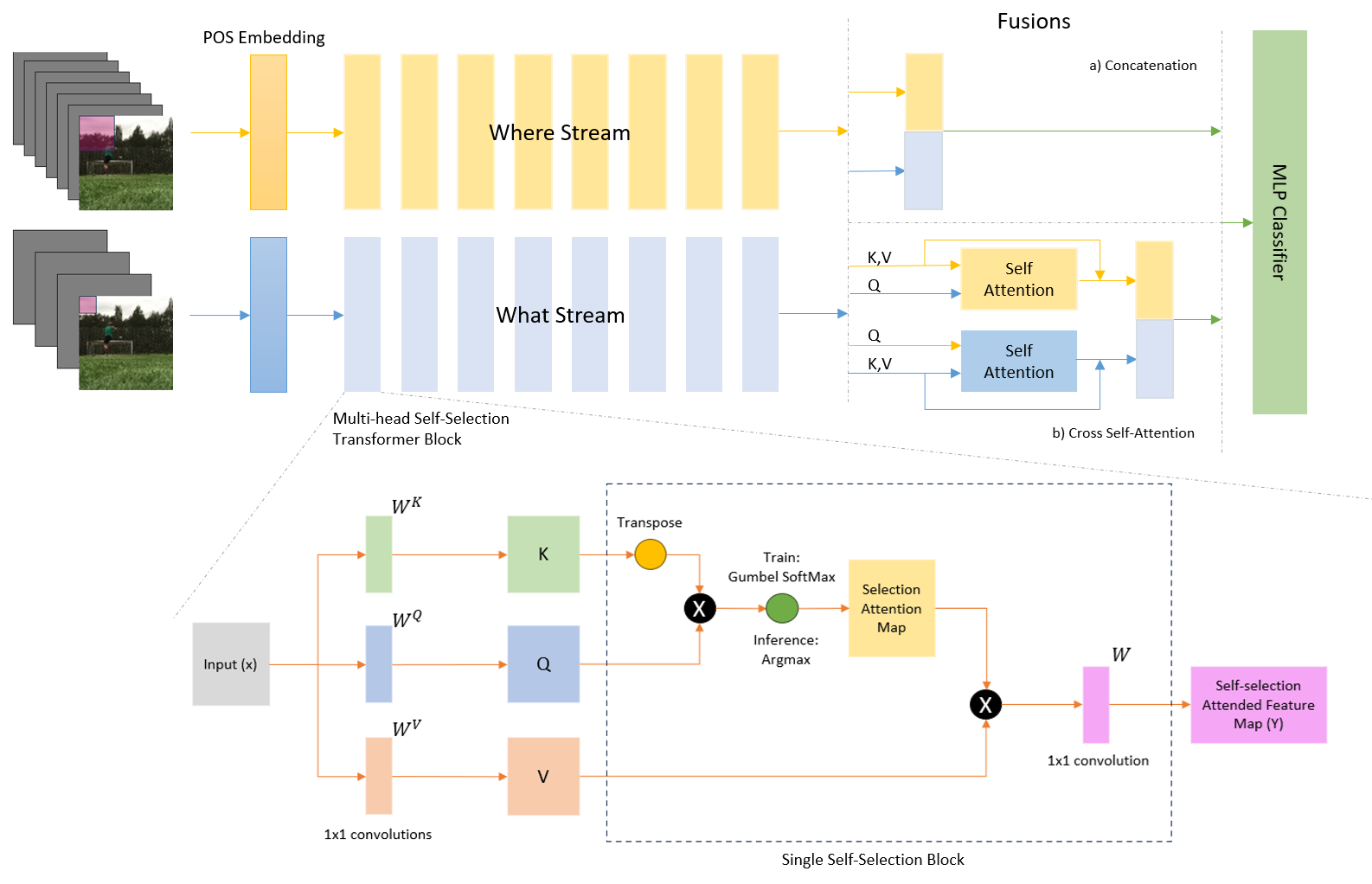}
    \caption{Overview of the neuro-aligned split-and-fuse architecture. The top panel shows the model split into a "what" stream that processes high-resolution frames at a lower frame rate and a "where" stream that processes lower-resolution frames at a higher frame rate to capture motion. The two streams are fused before the classifier via either simple feature concatenation or a cross-attention-based fusion block. The bottom panel illustrates the sparse self-selection module, where a winner-takes-all gate selects the most relevant token for each query, replacing standard dense self-attention.}\label{fig1}
\end{figure}

The Split-and-Fuse Transformer decomposes a baseline video transformer into two parallel streams that specialize in complementary aspects of the input (overview in Fig.~\ref{fig1})~\citep{goodale1992separate,Feichtenhofer2019}. The what stream processes frames with high resolution (patch size 16) at a reduced frame rate (8) to prioritize fine-grained spatial structure, whereas the where stream operates on frames with lower resolution (patch size 32) at an increased frame rate (32) to capture motion dynamics.

Each stream consists of an 8-layer transformer encoder derived from a TimeSformer backbone~\citep{Bertasius2021}, with separate patch embeddings, positional encodings, and classification tokens. After independent processing, the two final stream embeddings are merged by either (i) feature concatenation, or (ii) a lightweight cross-attention fusion block that performs bidirectional fusion between the two streams. 

Let $\mathbf{z}^{\mathrm{what}} \in \mathbb{R}^{D}$ and
$\mathbf{z}^{\mathrm{where}} \in \mathbb{R}^{D}$ denote the pooled
representations of the two streams, and let
$\mathbf{W}_{\mathrm{fuse}} \in \mathbb{R}^{D \times 2D}$ be a fusion
projection matrix. We define the concatenation fusion as
\begin{equation}
  \mathbf{z}^{\mathrm{fuse}} =
  \mathbf{W}_{\mathrm{fuse}}
  \begin{bmatrix}
    \mathbf{z}^{\mathrm{what}} \\
    \mathbf{z}^{\mathrm{where}}
  \end{bmatrix}.
  \label{eq:concat_fusion}
\end{equation}

In the cross-attention fusion block, first, the where stream attends to the what stream (where as queries, what as keys/values), and the resulting representations are added back to the what stream via a residual connection. Then, the what stream attends to the where stream (what as queries, where as keys/values), and the outputs are added back to the where stream. Finally, we concatenate the updated what and where representations before the classifier. Let $\mathbf{Z}^{\mathrm{what}}, \mathbf{Z}^{\mathrm{where}} \in \mathbb{R}^{L \times D}$ denote the token matrices from the what and where streams, respectively, and let
$\mathrm{CA}(\cdot,\cdot,\cdot)$ denote a multi-head cross-attention operator that takes queries, keys, and values. We define bidirectional cross-attention fusion as

\begin{equation}
  \tilde{\mathbf{Z}}^{\mathrm{what}}
  = \mathbf{Z}^{\mathrm{what}}
  + \mathrm{CA}\!\left(
      \mathbf{Z}^{\mathrm{where}},
      \mathbf{Z}^{\mathrm{what}},
      \mathbf{Z}^{\mathrm{what}}
    \right),
  \label{eq:ca_what}
\end{equation}
\begin{equation}
  \tilde{\mathbf{Z}}^{\mathrm{where}}
  = \mathbf{Z}^{\mathrm{where}}
  + \mathrm{CA}\!\left(
      \mathbf{Z}^{\mathrm{what}},
      \mathbf{Z}^{\mathrm{where}},
      \mathbf{Z}^{\mathrm{where}}
    \right),
  \label{eq:ca_where}
\end{equation}
and the final fused representation as
\begin{equation}
  \mathbf{Z}^{\mathrm{fuse}} =
  \begin{bmatrix}
    \tilde{\mathbf{Z}}^{\mathrm{what}} \\
    \tilde{\mathbf{Z}}^{\mathrm{where}}
  \end{bmatrix}.
  \label{eq:ca_fuse}
\end{equation}

\subsection{Self-Selection Block}

We propose a "winner-takes-all" modification to the canonical self-attention block, replacing the dense softmax weighting over values with a sparse, single-winner selection per query. Inspired by shift-based approximations in convolutional networks~\citep{lin2019tsm} and by competitive dynamics in biological vision~\citep{desimone1995neural}, our block computes the usual similarity scores between queries \(\mathbf{Q} \in \mathbb{R}^{B \times I \times D\ }\) and keys \(\mathbf{K} \in \mathbb{R}^{B \times J \times D\ }\), but instead of passing these through a softmax to produce a full attention map, we select the maximal key for each query channel and use that one value from \(\mathbf{V} \in \mathbb{R}^{B \times J \times D\ }\). In these notations, \(B\) denotes the batch size, \(I\) is the number of query positions, \(J\) is the number of key/value positions, and \(D\) denotes the number of dimensions in each query, key, or value vector, that is, the size of the feature-embedding space. During training, we approximate the non-differentiable argmax by a hard Gumbel-softmax~\citep{jang2017categorical, herrmann2020channel} with temperature \(\tau\), so that gradients still flow. At inference, we drop back to the exact argmax and a simple gather operation, as illustrated in Fig.~\ref{fig1}.

Given query, key, and value tensors
\begin{equation}
  \mathbf{Q} \in \mathbb{R}^{B \times I \times D}, \quad
  \mathbf{K} \in \mathbb{R}^{B \times J \times D}, \quad
  \mathbf{V} \in \mathbb{R}^{B \times J \times D},
  \label{eq1}
\end{equation}
we first form the raw similarity scores
\begin{equation}
  \mathbf{S}_{b,i,j} = \mathbf{Q}_{b,i,:} \cdot \mathbf{K}_{b,j,:},
  \label{eq2}
\end{equation}
where \(\mathbf{S} \in \mathbb{R}^{B \times I \times J}\).

In the standard transformer block~\citep{vaswani2017attention}, one would set
\begin{equation}
  \mathbf{A} = \operatorname{softmax}\left( \frac{\mathbf{S}}{\sqrt{d}} \right), \quad
  \mathbf{O} = \mathbf{A} \mathbf{V},
  \label{eq3}
\end{equation}
with \(\mathbf{A} \in \mathbb{R}^{B \times I \times J}\).

Instead, we define a selection map \(\mathbf{M} \in \{0,1\}^{B \times I \times J}\) that for each query index \((b,i)\) places a single 1 at the position of the maximal score,
\begin{equation}
  \mathbf{M}_{b,i,j} =
  \begin{cases}
    1, & j = \operatorname*{argmax}_{j'} \mathbf{S}_{b,i,j'}, \\
    0, & \text{otherwise},
  \end{cases}
  \label{eq4}
\end{equation}
and the block output is then computed by a gather:
\begin{equation}
  \mathbf{O}_{b,i,:}
  = \sum_{j=1}^{J} \mathbf{M}_{b,i,j} \, \mathbf{V}_{b,j,:}
  = \mathbf{V}_{b,\,\operatorname*{argmax}_{j} \mathbf{S}_{b,i,j},:}.
  \label{eq5}
\end{equation}

This single-winner approach removes the matrix multiplications that follow a dense attention map, reducing the memory footprint from \(O(BIJ)\) to \(O(BI)\) and lowering the compute from \(O(BIJD)\) to \(O(BID)\) in the gather step. However, a direct argmax is non-differentiable, preventing gradient flow. To address this, during training, we employ the Gumbel-softmax trick with hard sampling~\citep{jang2017categorical}:
\begin{equation}
  \mathbf{G}_{b,i,j}(\tau) =
  \frac{\exp\!\left( (\mathbf{S}_{b,i,j} + \mathbf{g}_{b,i,j}) / \tau \right)}
       {\sum_{j'} \exp\!\left( (\mathbf{S}_{b,i,j'} + \mathbf{g}_{b,i,j'}) / \tau \right)},
  \label{eq6}
\end{equation}
where \(\mathbf{g}_{b,i,j} \sim \mathrm{Gumbel}(0,1)\) are fresh noise samples, and \(\tau > 0\) is a temperature parameter. During each training forward pass, we assign
\begin{equation}
  \mathbf{A}_{b,i,j} = \operatorname{onehot}\!\left(
    \operatorname{argmax}_{j} \mathbf{G}_{b,i,j}
  \right),
  \label{eq7}
\end{equation}
and in backprop, we use
\begin{equation}
  \frac{\partial \mathbf{A}}{\partial \mathbf{S}} =
  \frac{\partial \mathbf{G}}{\partial \mathbf{S}}.
  \label{eq8}
\end{equation}

This hard Gumbel-softmax ensures that the forward output matches exactly the discrete selection map \(\mathbf{M}\), while the backward pass uses the continuous softmax gradient of \(\mathbf{G}\). By annealing \(\tau\) (for example, from 1.0 down toward 0.1), the distribution of \(\mathbf{G}\) increasingly concentrates on the actual maximum, thus matching the argmax more closely. We can compute
\begin{equation}
  \lim_{\tau \to 0^{+}} \mathbf{G}_{b,i,j}(\tau) =
  \begin{cases}
    1, & j = \operatorname*{argmax}_{j'} \left( \mathbf{S}_{b,i,j'} + \mathbf{g}_{b,i,j'} \right), \\
    0, & \text{otherwise}.
  \end{cases}
  \label{eq9}
\end{equation}

During training, the term \(\mathbf{g}_{b,i,j} \sim \mathrm{Gumbel}(0,1)\) introduces random noise to every score. At inference, we drop the noise and compute the selection mask \(\mathbf{M}\) by applying argmax directly, then use this mask to gather the corresponding value vectors as in Eq.~\eqref{eq5}. This yields deterministic routing at test time while retaining nonzero gradients for \(\mathbf{S}\) through the continuous relaxation during training.

\section{Experiments and results}

\subsection{Experimental setup}

All models are implemented in PyTorch and trained using mini-batches of fixed-length clips on NVIDIA RTX 4060 Ti GPUs. We consider different variants of our architecture: the what-only stream, the where-only stream, the "S What" stream which is the what-only stream with self-selection, the "S Where" stream which is the where-only stream with self-selection, the dual-stream WW-Former (without self-selection), and the dual-stream SWW-Former (with self-selection blocks in all transformer layers), each instantiated with either concatenation fusion or bidirectional cross-attention fusion. To stabilize sparse selection in the related models, the Gumbel-softmax temperature \(\tau\)~\citep{jang2017categorical} is initialized at a moderate value and annealed exponentially over training epochs, sharpening the winner-takes-all distribution.

For video recognition, we train and evaluate on Kinetics-400~\citep{Carreira2017} and Something-Something V2 (SSv2)~\citep{goyal2017something}. On Kinetics-400, all models are initialized from ImageNet-pretrained weights~\citep{deng2009imagenet}. For SSv2, we initialize from the corresponding Kinetics-400 checkpoints and fine-tune. In both datasets, videos are decoded into fixed-length clips, uniformly sampled over time, resized to a target spatial resolution, and normalized channel-wise; the what stream receives high-resolution, low-frame-rate clips, while the where stream receives low-resolution, high-frame-rate clips generated from the same raw video. We use identical data pipelines and augmentation strategies (including random cropping, horizontal flipping, and standard color jitter) across all models and baselines (including TimeSFormer, VideoSwin, MViT, MTV, and VideoMAE) to ensure a fair comparison. Classification performance is reported as top-1 and top-5 accuracy on the held-out validation set for each dataset.

We train all models with the AdamW optimizer, using a maximum learning rate of \(3\times 10^{-4}\), one cycle learning-rate decay, and a weight decay of 0.005. Each model is trained for 30 epochs on Kinetics-400 and 20 additional epochs on SSv2 with a global batch size of 64 clips (8 clips per GPU step on our NVIDIA RTX 4060 Ti setup). Input clips consist of 8 frames for the what stream and 32 frames for the where stream, sampled uniformly from each video. We use label smoothing with factor 0.05 and standard dropout in the transformer blocks; all hyperparameters are kept fixed across our model variants to isolate the effect of architectural changes. Table~\ref{tabsetup} summarizes the key setup details for all evaluated models, including parameter counts, input resolutions, and release years of baselines.

\begin{table}
\caption{Setup details for baseline and proposed models. The input size indicates frame count $\times$ spatial resolution (e.g., 8$\times$224$\times$224 for 8-frame, 224$\times$224 clips). For multi-stream models, largest frame count and spatial resolution is reported. Our models are derived from a 2021 TimeSformer backbone~\citep{Bertasius2021} and augmented with self-selection where indicated.}\label{tabsetup}
\begin{tabular*}{\textwidth}{@{\extracolsep\fill}lccc}
\toprule%
Model & Release Year & \# Params & Input Size \\
\midrule
I3D-R50~\citep{Carreira2017} & 2017 & 28.04 M & 8$\times$224$\times$224\\
SlowFast-R50~\citep{Feichtenhofer2019} & 2019 & 34.57 M & 32$\times$224$\times$224\\
SlowFast-R101~\citep{Feichtenhofer2019} & 2019 & 62.83 M & 32$\times$224$\times$224\\
CSN-R101~\citep{tran2019video} & 2019 & 22.21 M & 32$\times$224$\times$224 \\
TimeSFormer~\citep{Bertasius2021} & 2021 & 121.57 M & 8$\times$224$\times$224\\
TimeSFormer-HR~\citep{Bertasius2021} & 2021 & 122.02 M & 16$\times$448$\times$448\\
TimeSFormer-L~\citep{Bertasius2021} & 2021 & 121.61 M & 64$\times$224$\times$224\\
VideoSwin-B (V1)~\citep{Liu2022} & 2021 & 88.05 M & 32$\times$224$\times$224\\
MViT-B (V2)~\citep{Li2022} & 2022 & 36.61 M & 32$\times$224$\times$224\\
MTV-B~\citep{yan2022} & 2022 & 310 M & 32$\times$224$\times$224\\
VideoMAE (V2)~\citep{wang2023} & 2023 & 86.53 M & 16$\times$224$\times$224\\
\midrule
VideoSwin-G (V2)~\citep{liu2022swin} & 2022 & 3 B & 8$\times$320$\times$320\\
VideoPrism-G~\citep{zhao2024videoprism} & 2024 & 1 B & 8$\times$288$\times$288\\
InternVideo (V2)~\citep{wang2024internvideo2} & 2024 & 6 B & 16$\times$224$\times$224\\
FTP-UniFormer (V2)~\citep{lu2024enhancing} & 2024 & 620 M & 32$\times$224$\times$224\\
V-JEPA-G (V2)~\citep{assran2025v} & 2025 & 1 B & 16$\times$384$\times$384\\
\midrule
What Stream & 2021 & 57.75 M & 8$\times$224$\times$224\\
S What Stream & 2026 & 57.75 M & 8$\times$224$\times$224\\
Where Stream & 2021 & 83.07 M & 32$\times$224$\times$224\\
S Where Stream & 2026 & 83.07 M & 32$\times$224$\times$224\\
WW-Former Concat & 2026 & 141.44 M & 32$\times$224$\times$224\\
SWW-Former Concat & 2026 & 141.44 M & 32$\times$224$\times$224\\
WW-Former Attention & 2026 & 146.17 M & 32$\times$224$\times$224\\
SWW-Former Attention & 2026 & 146.17 M & 32$\times$224$\times$224\\
\bottomrule
\end{tabular*}
\end{table}

\subsection{Video recognition performance}

To assess the impact of the split-and-fuse design and the sparse self-selection blocks on recognition performance, we evaluate four main variants of our architecture: WW-Former and SWW-Former, each instantiated with either concatenation or bidirectional cross-attention fusion. For each variant, we also report results for the individual what-only and where-only streams. These models are compared against the original TimeSformer backbone~\citep{Bertasius2021} and strong representative video architectures, including MTV~\citep{yan2022}, VideoMAE~\citep{wang2023}, MViT~\citep{fan2021Multiscale,Li2022}, Video Swin~\citep{Liu2022,liu2022swin}, and SlowFast~\citep{Feichtenhofer2019}, on Kinetics-400 and SSv2~\citep{Carreira2017,goyal2017something}; the complete top-1 and top-5 accuracies are summarized in Table~\ref{tabacc}.

\begin{table}
\caption{Comparison with previous best results on video action classification datasets. Acc@N stands for Top N Accuracy. The models listed in the middle section were reported to be pre-trained on large-scale web data (other than ImageNet~\citep{deng2009imagenet} and Kinetics-400~\citep{Carreira2017} datasets), and their results are not directly comparable to other models which are listed.}\label{tabacc}
\begin{tabular*}{\textwidth}{@{\extracolsep\fill}lccccc}
\toprule%
& & \multicolumn{2}{@{}c@{}}{Kinetics-400} & \multicolumn{2}{@{}c@{}}{SS-V2} \\\cmidrule{3-4}\cmidrule{5-6}%
Model & \# Params & Acc@1 & Acc@5 & Acc@1 & Acc@5 \\
\midrule
I3D-R50~\citep{Carreira2017} & 28.04 M & 73.27 & 90.7 & 62.84 & -\\
SlowFast-R50~\citep{Feichtenhofer2019} & 34.57 M & 76.94 & 92.69  & 61.68  & 86.92\\
SlowFast-R101~\citep{Feichtenhofer2019} & 62.83 M & 77.9 & 93.27  & 63  & -\\
CSN-R101~\citep{tran2019video} & 22.21 M & 77 & 92.9  & -  & -\\
TimeSFormer~\citep{Bertasius2021} & 121.57 M & 77.9 & 93.2 & 59.1 & 85.6\\
TimeSFormer-HR~\citep{Bertasius2021} & 122.02 M & 79.6 & 94 & 61.8 & 86.9\\
TimeSFormer-L~\citep{Bertasius2021} & 121.61 M & 80.6 & 94.7 & 62 & 87.5\\
VideoSwin-B (V1)~\citep{Liu2022} & 88.05 M & 82.7 & 95.5  & 69.6  & 92.7\\
MViT-B (V2)~\citep{Li2022} & 36.61 M & 80.3 & 94.69  & 72.1  & -\\
MTV-B~\citep{yan2022} & 310 M & 81.8 & 95  & 67.6  & 90.1\\ 
VideoMAE (V2)~\citep{wang2023} & 86.53 M & 81.5 & 95.1 & 71.2 & 92\\
\midrule
VideoSwin-G (V2)~\citep{liu2022swin} & 3 B & 86.8 & - & - & -\\
VideoPrism-G~\citep{zhao2024videoprism} & 1 B & 87.2 & - & 68.5 & -\\
InternVideo-6B (V2)~\citep{wang2024internvideo2} & 6 B & 92.1 & - & 77.4 & -\\
FTP-UniFormer (V2)~\citep{lu2024enhancing} & 620 M & 93.4 & 99.3 & 79.8 & 98.9\\
V-JEPA-G (V2)~\citep{assran2025v}] & 1 B & 87.3 & - & 77.3 & -\\
\midrule
What Stream & 57.75 M & 72.31 & 90.01 & 55.61 & 85.18\\
S What Stream & 57.75 M & 77.48 & 93.54 & 58.48 & 86.34\\
Where Stream & 83.07 M & 72.46 & 90.61 & 58.45 & 86.01\\
S Where Stream & 83.07 M & 78.12 & 94.4 & 60.67 & 86.9\\
WW-Former Concat & 141.44 M & 78.91 & 93.86 & 68.03 & 90.92\\
SWW-Former Concat & 141.44 M & 81.5 & 95.58 & 69.93 & 92.31\\
WW-Former Attention & 146.17 M & 80.16 & 94.56 & 69.46 & 91.99\\
SWW-Former Attention & 146.17 M & 82.55 & 96.01 & 73.18 & 93.65\\
\bottomrule
\end{tabular*}
\end{table}

As reported in Table~\ref{tabacc}, the best-performing variant, SWW-Former with attention-based fusion, achieves 82.55\% and 73.18\% top-1 accuracy on Kinetics-400 and SSv2, respectively, outperforming similar-scale state-of-the-art models by more than 0.75 percentage points and 1 percentage point, excluding models with additional web-scale pretraining.

\subsection{Computational efficiency}

\begin{table}
\caption{Inference efficiency comparison (batch size 8 on RTX 4060 Ti, averaged over 100 runs). Inference time is in ms per batch; GFLOPs and peak CUDA memory (GB) are also reported. Self-selection variants (S) consistently reduce compute and latency relative to their dense-attention counterparts while improving accuracy.}\label{tabperf}
\begin{tabular*}{\textwidth}{@{\extracolsep\fill}lccc}
\toprule%
Model & Avg. Inference Time & GFLOPS & Peak CUDA Memory \\
\midrule
I3D-R50~\citep{Carreira2017} & 84.61 ms & 229.35 & 789.11 MB\\
SlowFast-R50~\citep{Feichtenhofer2019} & 274.78 ms & 406.84 & 1.06 GB\\
SlowFast-R101~\citep{Feichtenhofer2019} & 361.05 ms & 777.51 & 1.17 GB\\
CSN-R101~\citep{tran2019video} & 293.46 ms & 460.89 & 2.79 GB\\
TimeSFormer~\citep{Bertasius2021} & 386.73 ms & 1520.44 & 3.1 GB\\
TimeSFormer-HR~\citep{Bertasius2021} & 3932.16 ms & 2149.44 & 9.08 GB\\
TimeSFormer-L~\citep{Bertasius2021} & 3129.79 ms & 2160.33 & 5.96 GB\\
VideoSwin-B (V1)~\citep{Liu2022} & 1241.34 ms & 1310.9 & 6.98 GB\\
MViT-B (V2)~\citep{Li2022} & 1147.91 ms & 1216.16 & 5.8 GB\\
MTV-B~\citep{yan2022} & 2683.86 ms & 1453.01 & 5.98 GB\\
VideoMAE (V2)~\citep{wang2023} & 446.54 ms & 814.79 & 2.43 GB\\
\midrule
What Stream & 178.74 ms & 721.93 & 1.6 GB\\
S What Stream & 170.25 ms & 716.55 & 1.52 GB\\
Where Stream & 266.35 ms & 1041.93 & 2.07 GB\\
S Where Stream & 255.63 ms & 1020.15 & 2.01 GB\\
WW-Former Concat & 444.8 ms & 1763.86 & 3.33 GB\\
SWW-Former Concat & 423.43 ms & 1709.55 & 3.24 GB\\
WW-Former Attention & 444.94 ms & 1763.9 & 3.34 GB\\
SWW-Former Attention & 423.83 ms & 1710.33 & 3.27 GB\\
\bottomrule
\end{tabular*}
\end{table}

Table~\ref{tabperf} reports average inference time, GFLOPs, and peak GPU memory usage for our dual-stream models and comparable transformer baselines, measured for a batch size of 8 on an RTX 4060 Ti and averaged over 100 runs. 

\begin{figure}
\centering
\includegraphics[width=0.9\textwidth]{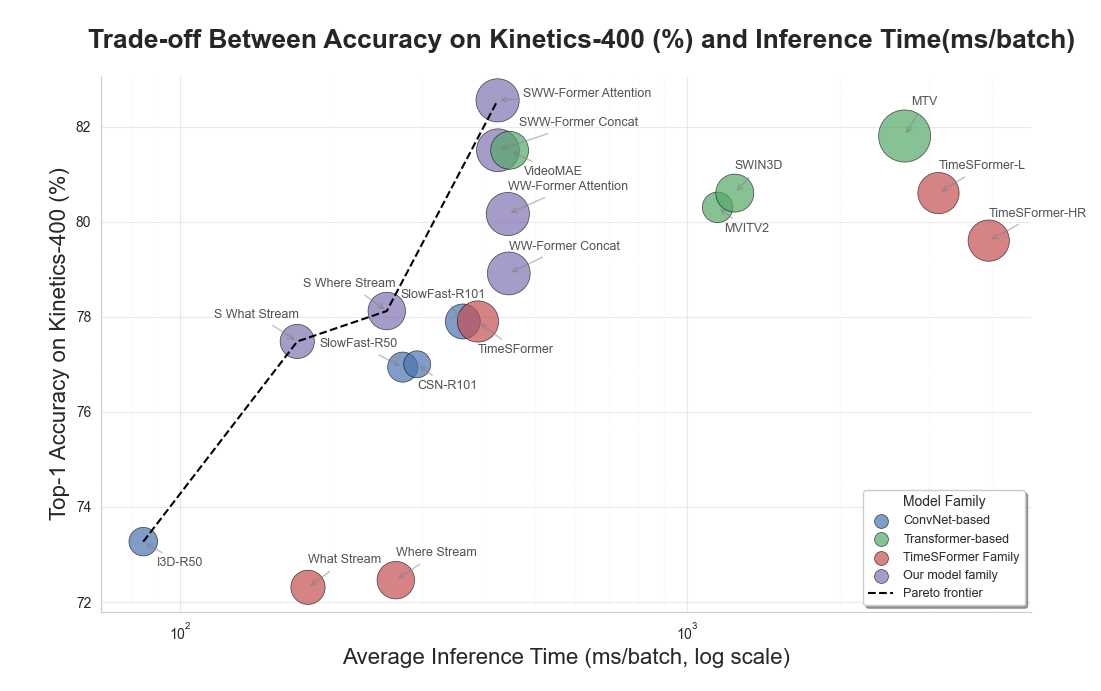}
\caption{Trade-off between top-1 accuracy and inference time (ms per batch) on Kinetics-400~\citep{Carreira2017} for our models and others of similar size and pretraining. Models in the top-left region are both more accurate and faster. Bubble size indicates the number of trainable parameters.}\label{fig2}
\end{figure}

\begin{figure}
\centering
\includegraphics[width=0.9\textwidth]{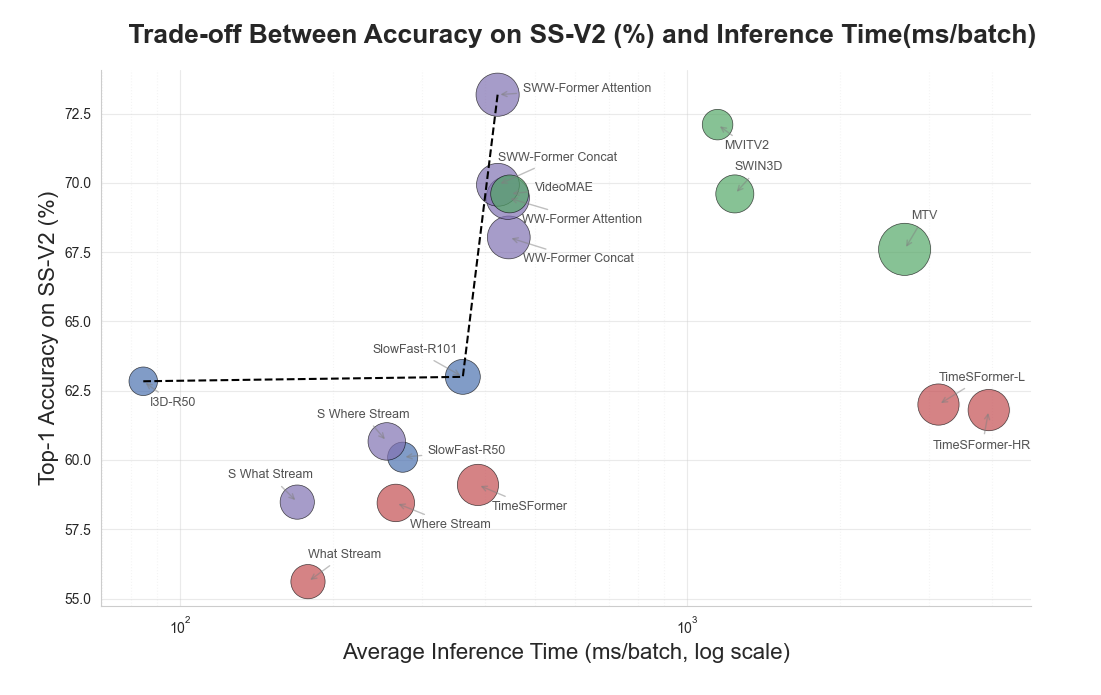}
\caption{Trade-off between top-1 accuracy and inference time (ms per batch) on SSv2~\citep{goyal2017something} for our models and others of similar size and pretraining. Models in the top-left region are both more accurate and faster. Bubble size indicates the number of trainable parameters.}\label{fig3}
\end{figure}

Self-selection blocks yield consistent efficiency gains across both fusion strategies. For attention-fusion variants, WW-Former requires 444.94~ms per batch, 1763.90 GFLOPs, and 3.34~GB peak memory, whereas SWW-Former with self-selection reduces these by 4.7\% to 423.83~ms, 3.0\% to 1710.33 GFLOPs, and 2.1\% to 3.27~GB while boosting top-1 accuracy (80.16\% to 82.55\% on Kinetics-400; 69.46\% to 73.18\% on SSv2; see Tables~\ref{tabsetup} and~\ref{tabacc}). A similar pattern holds for concatenation-fusion variants. In the accuracy-latency plane (Fig.~\ref{fig2} and Fig.~\ref{fig3}), SWW-Former variants occupy the Pareto frontier among models of comparable-scale and pretraining (no other databases for pretraining besides ImageNet~\citep{deng2009imagenet} and Kinetics-400~\citep{Carreira2017}).

While concatenation fusion minimizes additional compute (141.44M params, 423.43~ms inference for SWW-Former Concat), cross-attention fusion adds modest overhead (146.17M params, +0.4~ms) but remains efficient due to lightweight bidirectional design. Both strategies benefit from self-selection sparsity, with attention fusion preserving Pareto dominance despite extra parameters.

\subsection{Gumbel-softmax temperature ablation}

The temperature parameter \(\tau\) in the hard Gumbel-softmax~\citep{jang2017categorical} controls how sharply the selection distribution concentrates on the highest-scoring key for each query in the self-selection block. During training, larger \(\tau\) yields a softer, more exploratory distribution over keys, while smaller \(\tau\) makes the sampling behave more like a hard argmax, approximating the deterministic inference-time behavior. In the main experiments, \(\tau\) is annealed from a moderate initial value toward a lower final value, allowing early training to benefit from smoother gradients while late training encourages commitment to single winners, similar in spirit to prior work that uses Gumbel-softmax for differentiable channel selection~\citep{herrmann2020channel}.

To assess the sensitivity of our model to the choice of Gumbel-softmax temperature \(\tau\), we trained SWW-Former variants with fixed \(\tau\) values ranging from 0.1 to 10 and measured mean top-1 accuracy and standard deviation over 5 runs on the Kinetics-400 validation set.

\begin{figure}
\centering
\includegraphics[width=0.7\textwidth]{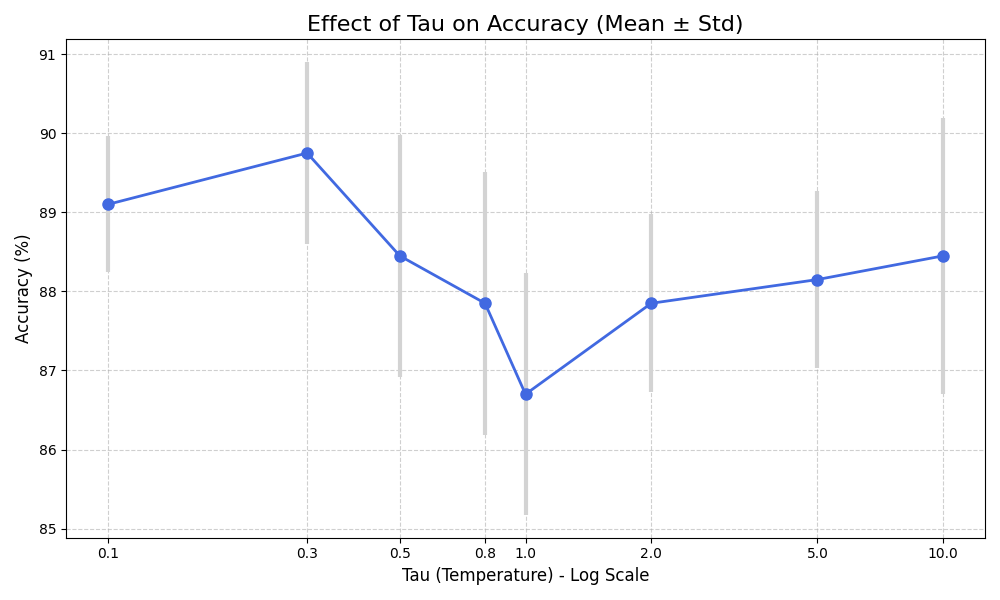}
\caption{Effect of fixed Gumbel-softmax temperature \(\tau\) on top-1 validation accuracy of SWW-Former (Attention), with error bars indicating one standard deviation across 5 runs. Accuracy varies by less than about two percentage points across the tested range and peaks around \(\tau = 0.3\).}\label{fig4}
\end{figure}

As shown in Fig.~\ref{fig4}, performance is relatively stable across the tested range. As a result all temperatures achieve accuracy within roughly two percentage points of each other, indicating that the self-selection mechanism is robust to the exact choice of \(\tau\).  The highest mean accuracy is observed around \(\tau = 0.3\), with \(\tau = 0.1\) and larger values (e.g., \(\tau \geq 5\)) yielding slightly lower but still comparable performance.  These results suggest that using a moderately low temperature provides a good practical default without requiring fine-tuning of \(\tau\).

\subsection{Invariance and robustness analysis}

To evaluate generalization beyond clean validation data, we assessed model robustness by applying random spatial and photometric perturbations at test time, following prior work on invariances and input-variance robustness~\citep{tschannen2020self, fadaei2024}, measuring the relative drop in top-1 accuracy averaged over 5 trials on the Kinetics-400~\citep{Carreira2017} and Something-Something V2~\citep{goyal2017something} validation sets. Specifically, we applied random spatial augmentations including color jittering, flipping, reversing, rotation, cropping, and adjustments to brightness, contrast, and hue. Detailed application of these augmentations are included in our shared codebase.

\begin{figure}
\centering
\includegraphics[width=1\textwidth]{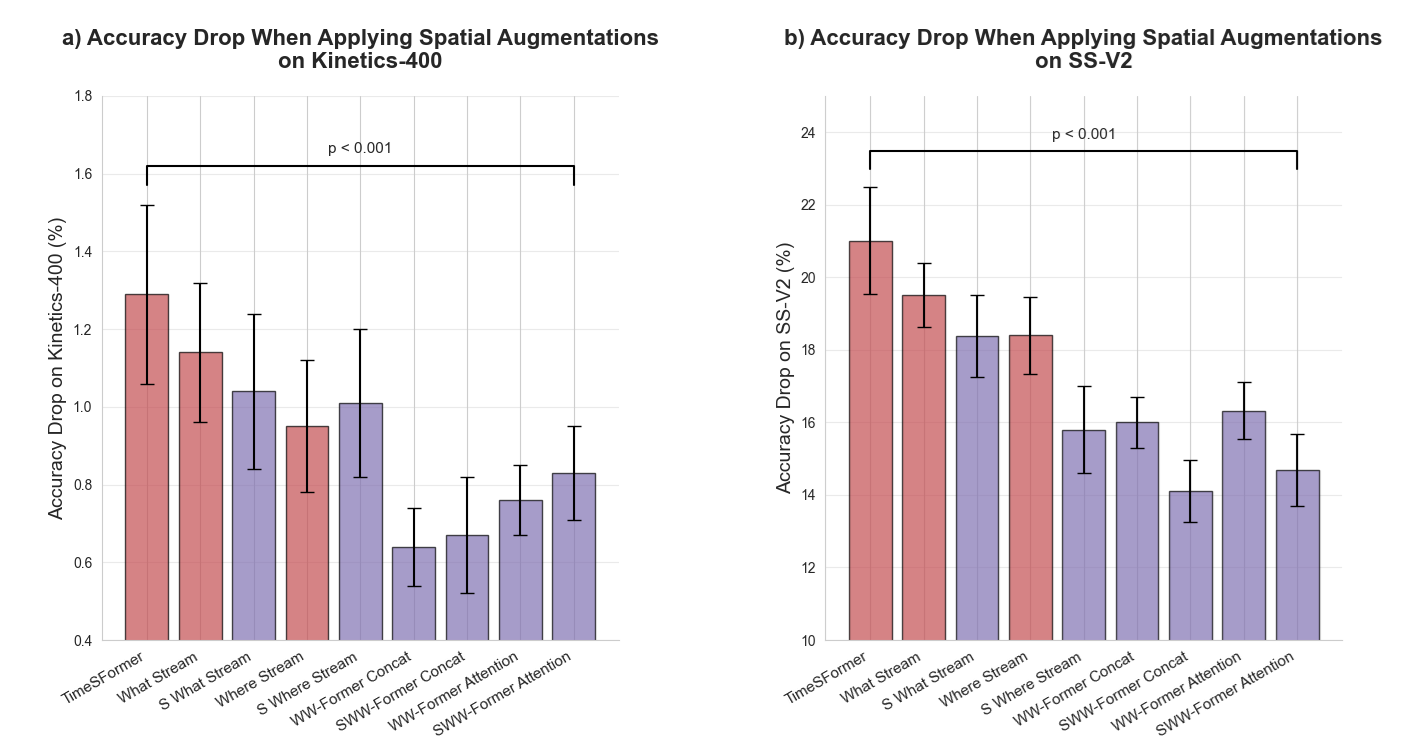}
\caption{Percentage drop in top-1 accuracy under spatial augmentations on (a) Kinetics-400~\citep{Carreira2017} and (b) Something-Something V2~\citep{goyal2017something}. The error bars report the standard deviation of 5 trials and p-values test differences from TimeSformer and are calculated by bootstrap test (N=100,000).}\label{fig5}
\end{figure}

Table~\ref{tabinvariance} and Fig.~\ref{fig5} show the resulting drop in accuracy. Our proposed multi-stream models  are more resilient to these changes, with a smaller drop in accuracy compared to the TimeSFormer baseline (p-value < 0.01; bootstrap tests with N=100000). Larger accuracy drops on SSv2 (\textasciitilde 14-21\%) versus Kinetics (\textasciitilde 0.6-1.3\%) reflect dataset differences. SSv2 emphasizes fine-grained temporal ordering and human-object interactions sensitive to spatial augmentations like rotation/cropping, while Kinetics-400 focuses on broader action recognition more invariant to such perturbations.

\begin{table}[h]
\caption{Reported top-1 accuracy and the calculated accuracy drop without and with augmentations in Kinetics-400~\citep{Carreira2017} and Something-Something V2~\citep{goyal2017something} Datasets.}\label{tabinvariance}
\begin{tabular*}{\textwidth}{@{\extracolsep\fill}lcccccc}
\toprule%
& \multicolumn{3}{@{}c@{}}{Kinetics-400} & \multicolumn{3}{@{}c@{}}{SS-V2} \\\cmidrule{2-4}\cmidrule{5-7}%
Model & Normal & Augmented & Drop & Normal & Augmented & Drop \\
\midrule
TimeSFormer~\citep{Bertasius2021} & 77.9 & 76.61 & 1.29 & 59.1 & 38.09 & 21.01 \\
What Stream & 72.31 & 71.17 & 1.14 & 55.61 & 36.09 & 19.52\\
S What Stream & 77.48 & 76.44 & 1.04 & 58.48 & 40.1 & 18.38\\
Where Stream & 72.46 & 71.51 & 0.95 & 58.45 & 40.05 & 18.4\\
S Where Stream & 78.12 & 77.11 & 1.01 & 60.67 & 44.87 & 15.8\\
WW-Former Concat & 78.91 & 78.27 & 0.64 & 68.03 & 52.03 & 16\\
SWW-Former Concat & 81.5 & 80.83 & 0.67 & 69.93 & 55.83 & 14.1\\
WW-Former Attention & 80.16 & 79.4 & 0.76 & 69.46 & 53.14 & 16.32\\
SWW-Former Attention & 82.55 & 81.72 & 0.83 & 73.18 & 58.49 & 14.69\\
\bottomrule
\end{tabular*}
\end{table}

\subsection{EEG representational similarity analysis}

To assess whether our split-and-fuse transformers mirror human brain dynamics, and following prior model–brain alignment work~\citep{yamins2016,schrimpf2018,conwell2024large,celeghin2023}, we recorded 128-channel scalp EEG (g.HIamp/g.GAMMASys, 10-10 layout plus interleaved) at 512 Hz (downsampled to 200~Hz) while participants viewed selected clips from the HVU dataset~\citep{Diba2020}. Signals were high-pass filtered (1~Hz), notch-filtered (49-51 Hz), re-referenced to the channel average, epoched (-200~ms to 3.4~s post-onset; -200 ms to 400 ms for images), baseline-corrected, and cleaned of artifacts via visual inspection and ICA.

Thirty healthy adults (16~M/14~F, ages 20 to 30) from a pre-screened pool gave written informed consent under an IRB-approved protocol compliant with the Code of Ethics. Exclusion criteria included neurological disorders, uncorrected vision deficits, epilepsy, and head injury; a general practitioner verified eligibility. We handled the personal data in accordance with institutional policy. Each participant was compensated over the national minimum wage to cover time and travel expenses. We included the participant instructions, task screenshots, and consent form in the appendix~\ref{appendixb}. The EEG data can be made available upon reasonable request to the corresponding author, but will not be made publicly available due to privacy concerns, institutional restrictions, and other considerations.

We selected 252 intact 3 s clips from eight categories (Animal, Art, Food, Garden, Kid, Music, Sports, and Vehicle) and 21 subcategories from the HVU dataset~\citep{Diba2020}; from which we randomly scrambled 84 videos by frame shuffling, and we have shown their middle frame as 50 ms static images as separate stimuli. Stimuli subtended a 5° visual angle, ran at 60 Hz on a gray background, and appeared in three trial types: coherent video, scrambled video, and static image. Each two-hour session comprised 336 video trials and 252 image trials (each repeated four times), with 400 ms inter-trial intervals, pseudo-random fixation-cross color changes for attention checks, and rest breaks every \textasciitilde 15 minutes.

\begin{figure}
\centering
\includegraphics[width=1\textwidth]{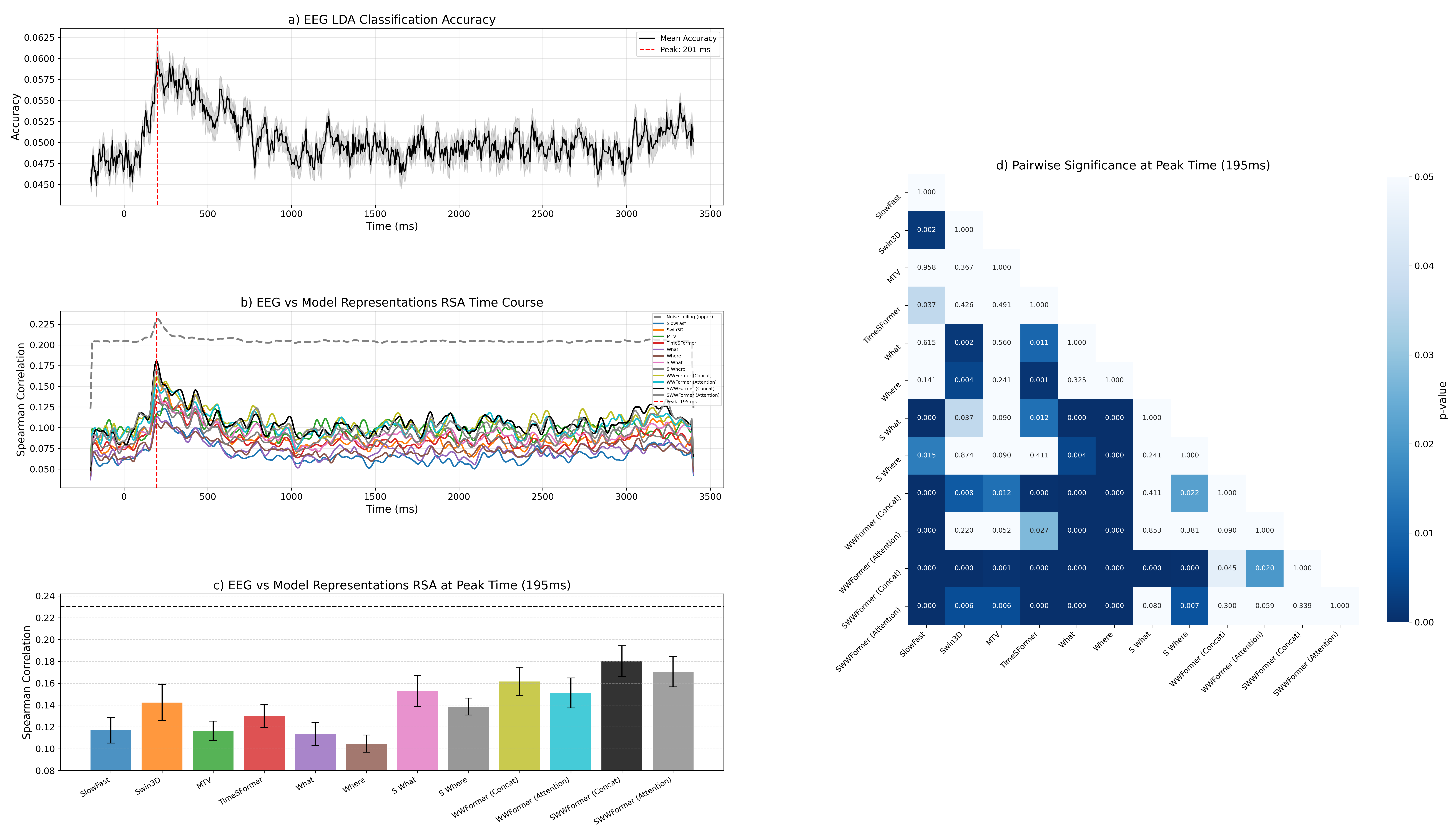}
\caption{Model-EEG alignment. (a) Time-resolved multiclass LDA decoding accuracy for 21 visual subcategories. (b) Time-resolved mean Spearman RSA correlations between model RDMs and EEG RDMs with noise ceiling. (c) Peak mean RSA correlations (195 ms) across models (The error bars indicate SEM), with noise ceiling. (d) Pairwise Wilcoxon signed-rank tests on peak RSA scores.}\label{fig6}
\end{figure}

For model comparison, we first applied multi-class linear discriminant analysis (LDA)~\citep{fisher1936} at each time point to decode the 21 stimulus subcategories from the EEG, providing a simple baseline alongside more complex EEG decoding approaches~\citep{schirrmeister2017}. This revealed a transient accuracy peak at approximately 200 ms after video onset, indicative of early category-selective visual processing consistent with prior MEG/EEG studies~\citep{Cichy2014}(Fig.~\ref{fig6} a). 

To quantify the alignment between model and human EEG representations, we performed time-resolved Representational Similarity Analysis (RSA)~\citep{kriegeskorte2008}. For each model, the pairwise dissimilarities of its video embeddings were computed and flattened into Representational Dissimilarity Matrices (RDMs). EEG RDMs were constructed for each participant at each time point from scalp-recorded responses, and flattened analogously. Spearman correlation coefficients were then computed between the model and EEG RDMs at each time point, yielding participant-level time courses of representational similarity. To contextualize these correlations, we estimated a noise ceiling reflecting the maximum explainable variance given inter-subject reliability. Specifically, for each time point, the upper bound was calculated as the mean Spearman correlation between each participant’s EEG RDM and the group-average RDM including that participant. All correlations were Fisher-Z transformed before averaging and smoothed with a five-timepoint moving average to reduce noise. The resulting noise ceiling provides a reference for the best achievable model-EEG alignment in our dataset, enabling normalization of model correlations as a fraction of explainable variance.

This yielded brain-model correspondence time courses for each model, all peaking around 195 ms and mirroring the LDA profile (Fig.~\ref{fig6} b). To directly compare models, we examined RSA correlations at the 195 ms peak (Fig.~\ref{fig6} c). We conducted two-sided pairwise Wilcoxon signed-rank tests to assess the statistical significance of differences between model pairs (Fig~\ref{fig6} d). Our SWW-Former model with concatenation fusion achieves 0.18 RSA correlation, which, excluding the SWW-Former with attention fusion, is significantly higher than all tested models (p < 0.05 two-sided Wilcoxon signed-rank test against all other baselines) and reached up to 78\% of the estimated noise ceiling at the peak time, indicating substantial capture of neural representational structure.

The what and where streams show closely similar correlation values of 0.11 and 0.10, respectively (p-value = 0.325; two-sided pairwise Wilcoxon signed-rank test), indicating that both streams contribute similarly to the model and its neuro-alignment. Furthermore, the Concatenation and Attention fusion variants of WW-Former, with multi-stream, and S What and S Where stream models, which implement self-selection blocks instead of self-attention blocks, achieve 0.16, 0.15, 0.15 and 0.14 correlation, respectively. These results show significant improvement over the standard What and Where streams (For every comparisons of these variants with each of What and Where streams, p-value \textless  0.005; two-sided pairwise Wilcoxon signed-rank test), demonstrating that each of our introduced contributions helps with the neural alignment of the model. SWW-Former with concatenation achieves 0.18 correlation comparing to 0.17 correlation by SWW-Former attention (P-value \textgreater  0.33; two-sided paired Wilcoxon signed-rank test) which suggests our different fusion mechanics have no significant effect on brain-model alignment.

\begin{figure}
\centering
\includegraphics[width=1\textwidth]{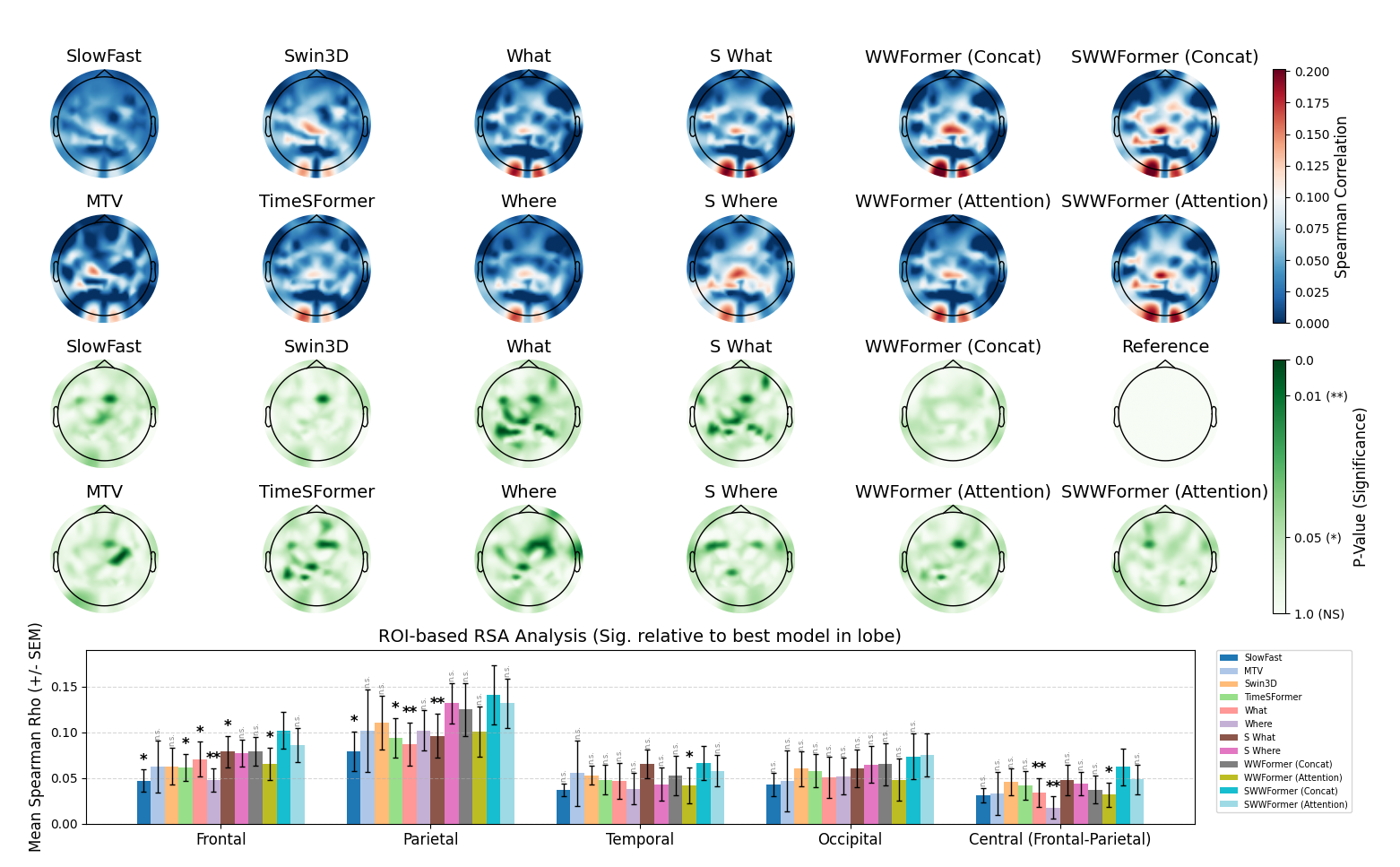}
\caption{Channel-wise and regional RSA between video models and EEG at peak latency (195 ms). The top two rows show scalp topographies of Spearman correlations. The middle two rows show significance maps from paired t-tests against best model per lobe where darker green indicates lower p-values and significant difference from SWW-Former with concatenation fusion. The plot shows mean RSA by ROI including frontal, parietal, temporal, occipital, central (fronto-parietal) regions, with markers for two-sided paired t-tests against best model per ROI. The error bars in this plot indicate SEM.}\label{fig7}
\end{figure}

Finally, we examined the spatial organization of brain-model correspondence by mapping channel-wise RSA onto the scalp. We computed channel-wise significance tests relative to the best-aligned architecture and aggregated correlations within canonical lobar ROIs. This analysis, summarized in Fig.~\ref{fig7}, shows that the SWW-Former variants yield higher, more spatially structured correlations in parietal, lateral temporal, and other regions than monolithic baselines. The SWW-Former recorded average correlations of 0.10, 0.14, 0.07, 0.07, and 0.06 for frontal, parietal, temporal, occipital, and central region electrodes, respectively. The two-sided paired t-tests with SWW-Former variants are reported in Fig.~\ref{fig7}. This analysis shows that the fused sparse model is consistently among the top performers across various lobes.

Furthermore, Fig.~\ref{fig7} ROI analysis confirms that the what/S What streams align better with temporal electrodes (ventral-like), while Where/S Where streams align better with parietal (dorsal-like), consistent with classic dorsal–ventral dissociations~\citep{goodale1992separate,creem2001} and recent DVSD modeling~\citep{Zareh2024} (p \textless 0.05, paired t-tests against the counterpart stream for each comparison). We also included layer-wise RSA analysis of the S What and S Where streams in the appendix~\ref{appendixa}, detailing how different layers of our streams correlate with the recorded data.

\section{Discussion}

This work presented a Neuro-Aligned Split-and-Fuse Transformer. It decomposes video processing into specialized "what" and "where" streams and augments them with a sparse self-selection mechanism. Across Kinetics-400 and Something-Something V2, the split-and-fuse design with self-selection achieved competitive or state-of-the-art performance among models of comparable scale without relying on massive web pre-training. Our model also operates at the Pareto frontier of accuracy versus inference time in both datasets. These architectural changes yielded representations that are more closely correlated with human EEG responses than those from conventional monolithic transformers, suggesting that explicit pathway separation and winner-takes-all selection are beneficial for both engineering and NeuroAI, and contribute to ongoing efforts to align high-level visual representations in brains and machines~\citep{conwell2024large, celeghin2023, soni2024conclusions} and in line with classic dorsal–ventral dissociations and competitive-attention mechanisms in vision~\citep{goodale1992separate, creem2001, desimone1995neural, itti2001computational}.

Beyond summarizing empirical gains, a central claim of this paper is mechanistic: sparse competitive selection together with pathway specialization constrains the model to forward only a small subset of informative states, echoing sparse and dynamic selection mechanisms in modern neural architectures~\citep{shazeer2017, beltagy2020, kitaev2020, Rao2021}, which both reduces computation and produces a representational geometry that more closely resembles early/mid-latency cortical responses reported in human neuroimaging~\citep{Cichy2014,yamins2016,schrimpf2018}. Our analyses are consistent with this account in several ways. Replacing dense attention by single-winner selection reduces latency markedly while largely preserving top-line accuracy and robustness (Fig.~\ref{fig2} and Fig.~\ref{fig3}). Furthermore, the RSA results (Fig.~\ref{fig6} and Fig.~\ref{fig7}) shows that whole and partial brain-model alignments increases in models with self selection.

That said, the neural-alignment results require careful qualification. The peak Spearman correlation we report for the SWW-Former (0.18) is modest in absolute magnitude, reflecting a small but reliable fraction of between-subject variance in scalp EEG. To contextualize this effect, we computed a noise ceiling, representing the best-case model-EEG correlation given inter-subject reliability in our dataset~\citep{kriegeskorte2008}. At the time of peak correlation (195 ms), the SWW-Former explains approximately 78\% of the explainable variance, based on the mean noise ceiling (correlation = 0.23). Reporting correlations alongside noise-ceiling fractions makes clear that our models capture neural responses more closely than baseline transformers, while substantial neural variance remains unexplained. 

Several alternative explanations could inflate model-brain similarity, and we performed control analyses to address them. Time-shifted RSA (random circular temporal shifts of EEG relative to stimuli) yields correlations near zero, ruling out trivial temporal autocorrelation as the sole driver. Furthermore, we computed RDMs from other convolutional and transformer based models such as SlowFast~\citep{Feichtenhofer2019}, Swin3D~\citep{Liu2022}, MTV~\citep{yan2022} and TimeSFormer~\citep{Bertasius2021}, which produced substantially lower correlations than our models (Fig.~\ref{fig6}).

We further examined the spatial organization of model-brain correspondence by mapping channel-wise RSA across the scalp and aggregating correlations within canonical frontal, parietal, temporal, occipital and central regions (Fig.~\ref{fig7}). This analysis revealed that our split-and-fuse models produce more structured and generally higher correlations over occipital, parietal and lateral temporal electrodes than monolithic transformer baselines, indicating that sparse multi-stream processing yields representations that better capture the distributed geometry of human visual responses. Moreover, the what and where streams show complementary alignment profiles, with the what stream more strongly correlated in temporal regions and the where stream more strongly correlated in parietal regions, consistent with the functional dissociation of ventral and dorsal visual pathways~\citep{goodale1992separate, creem2001}. Together, these spatially resolved EEG results extend our findings beyond simple decoding accuracy or global similarity scores, and position our framework as a richer, temporally and regionally specific testbed for NeuroAI hypotheses about sparse competition, pathway specialization and their roles in shaping efficient visual representations.

Methodological limitations must be acknowledged. First, scalp EEG provides limited spatial precision and mixes cortical sources; thus channel-wise topographies must be interpreted cautiously. Source-reconstruction or higher-resolution modalities (MEG, fMRI, or intracranial recordings) could more precisely test cortical mapping claims. Second, our sample size (N = 30) allows for reliable estimation of group-level RSA correlations. While individual participants' correlations are modest in magnitude (0.1 to 0.18), averaging across participants yields stable time courses, and the time-resolved noise ceiling contextualizes these effects relative to the maximum explainable variance in our dataset. Thus, the observed neural alignment is robust at the group level, even though single-participant correlations remain small. However, larger samples and more diverse stimuli would strengthen claims about generality. Third, the selection mechanism was evaluated only in a TimeSFormer-derived backbone and two fusion strategies; whether comparable gains persist across other transformer families or large-scale pretraining regimes remains to be tested.

These caveats suggest several concrete follow-up steps. One direction is to extend the neuro-alignment analysis to fMRI, MEG, or intracranial recordings and to additional datasets, to separate genuine cortical principles from data-specific regularities. Another is to apply sparse What-Where transformers to temporal localization, video-language grounding, and online decision-making, and ask whether the efficiency gains we observe here actually improve behavior in interactive or resource-constrained settings. A longer-term avenue is to embed self-selection into scalable video backbones and multimodal foundation models, aiming to jointly improve data efficiency, computational cost, and alignment with human perception and cognition.

\section{Conclusion}

In this work, we introduced neuro-aligned split-and-fuse video transformers with sparse winner-takes-all self-selection, achieving competitive or SOTA accuracy (82.55\% on Kinetics-400~\citep{Carreira2017}, 73.18\% on SSv2~\citep{goyal2017something}), inference Pareto efficiency, robustness, and EEG alignment ($\rho$ = 0.18, 78\% of the noise ceiling~\citep{kriegeskorte2008}) superior to baselines. In sum, our results suggest that pathway specialization and sparse competition are promising inductive biases for efficient, neuro-aligned video models. While the gains in brain-model correspondence are modest, they are consistent, spatially structured, and robust to several controls, supporting further exploration of biologically inspired sparsity in transformer architectures and paving the way for NeuroAI in resource-constrained vision tasks.









\printcredits

\section{Declaration of generative AI and AI-assisted technologies in the manuscript preparation process}

During the preparation of this work the author(s) used AI-based language editing tools solely to improve the grammar and readability of the manuscript. After using this tool/service, the authors reviewed and edited the content as needed and take full responsibility for the content of the published article.

\section{Declaration of competing interest}

The authors declare that they have no known competing financial interests or personal relationships that could have appeared to influence the work reported in this paper.

\section{Data and Code availability}

The video datasets used in this study, including Kinetics-400, Something-Something V2 and the HVU dataset, are available from the previously published works cited in this manuscript. EEG data collected for this study are not publicly available due to institutional regulations and participant privacy considerations, but are available from the corresponding author upon reasonable request.

The research codebase, trained models, and all related codes for various research and tests conducted in this manuscript are available in our public github repository: https://github.com/amfad33/SWW-Former

\section{Acknowledgement}

The authors gratefully acknowledge the support provided by the Iran National Science Foundation (INSF), which funded this work under project No. 4032942. This support has been instrumental in advancing our research objectives and enabling the successful completion of this study. We extend our appreciation to INSF for its commitment to fostering scientific innovation and excellence. We gratefully acknowledge the support of the National Brain Mapping Laboratory (NBML) for their assistance in EEG data acquisition for this project and for providing access to their facility. All procedures involving human participants were conducted in accordance with guidelines and relevant ethical codes, and were approved by the institutional ethics committee. 

\appendix

\section{Appendix: Layer-wise RSA for the Streams}\label{appendixa}

\begin{figure}
\centering
\includegraphics[width=0.9\textwidth]{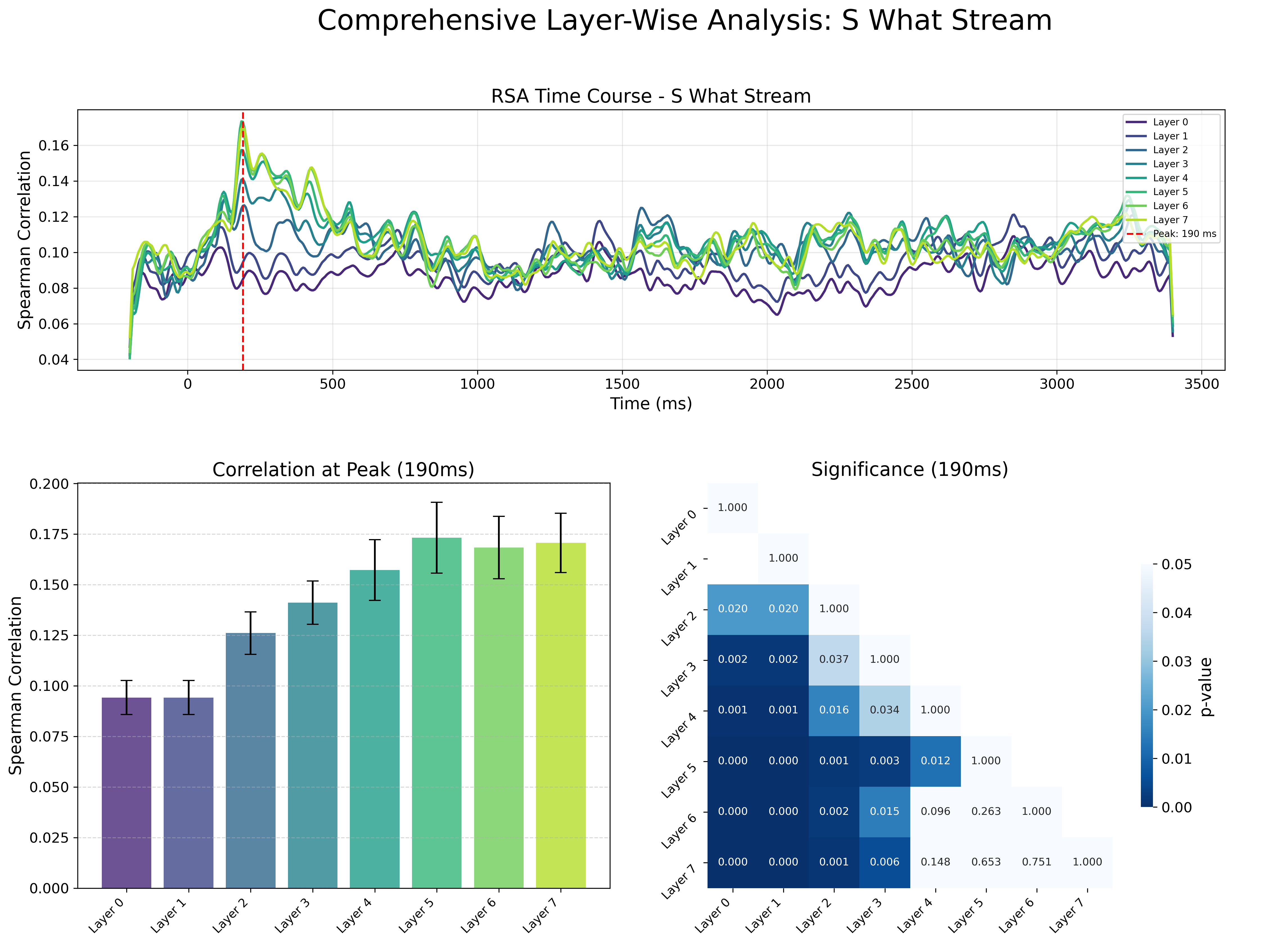}
\caption{Layer-wise S What stream-EEG RSA analysis. Layer-wise RSA in the S What stream peaks at around 190 ms, with higher layers exhibiting the strongest correlations.}\label{figs1}
\end{figure}

\begin{figure}
\centering
\includegraphics[width=0.9\textwidth]{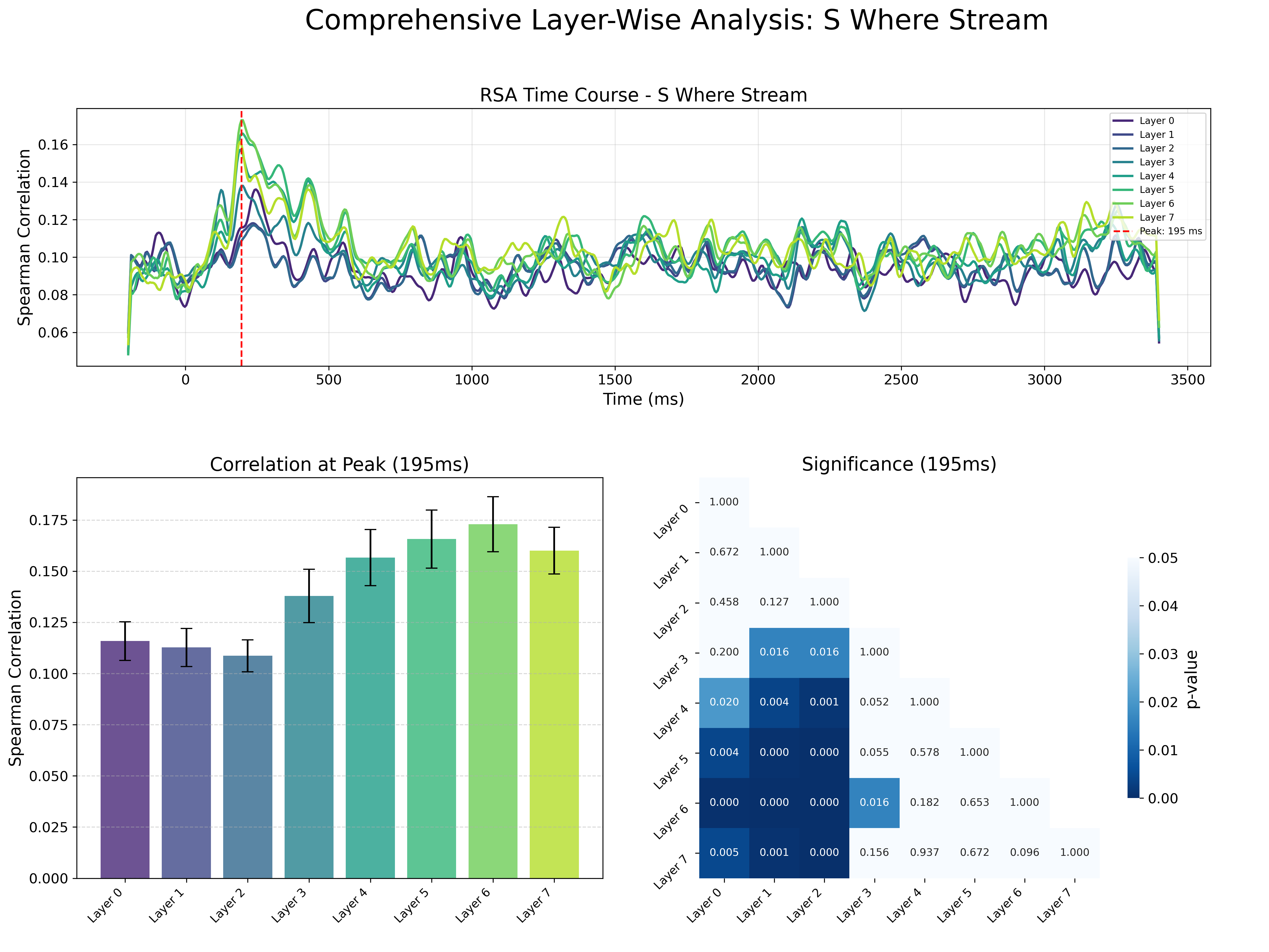}
\caption{Layer-wise S Where stream-EEG RSA analysis. Layer-wise RSA reveals that deeper S Where layers show progressively stronger alignment with EEG around 195 ms.}\label{figs2}
\end{figure}

The layer-wise RSA analyses for the S What and S Where streams (Fig.~\ref{figs1} and Fig.~\ref{figs2}) show that neural alignment increases monotonically with network depth, with the highest correlations emerging in the final layers around the early categorization peak in the EEG (approximately 190–195 ms), consistent with time-resolved RSA and object recognition dynamics reported in prior work~\citep{kriegeskorte2008, Cichy2014}. This pattern indicates that deeper representations in both streams capture more abstract, temporally precise information that more closely matches the structure of human visual responses during naturalistic video viewing, supporting the idea that the self-selection blocks preserve, and in fact enhance, brain-like representational geometry as information flows through the model~\citep{yamins2016,schrimpf2018}. The time-course plots in each figure demonstrate that all layers exhibit a clear transient peak near the behavioral decoding maximum, suggesting a consistent latency of category-selective processing across the hierarchy, while the bar plots at the peak make explicit that late layers dominate the overall brain-model correspondence.

The significance matrices further show that differences between early and late layers are statistically reliable (pairwise Wilcoxon signed-rank tests) at the peak latency, particularly between the shallowest and deepest layers in each stream. In the S Where stream, higher layers (5–7) exhibit significantly stronger RSA than the lower ones, reflecting the accumulation of motion and action-relevant features that align with dorsal, spatial processing. We observe a similar progression in the S What stream, but with slightly earlier peak timing and stronger correlations in ventral-leaning regions, consistent with object and appearance-based coding. Together, these statistics show that the sparse selection mechanism not only maintains alignment but also sharpens it in later stages, when the model’s internal representations are closest to human EEG.

\begin{figure}
\centering
\includegraphics[width=0.9\textwidth]{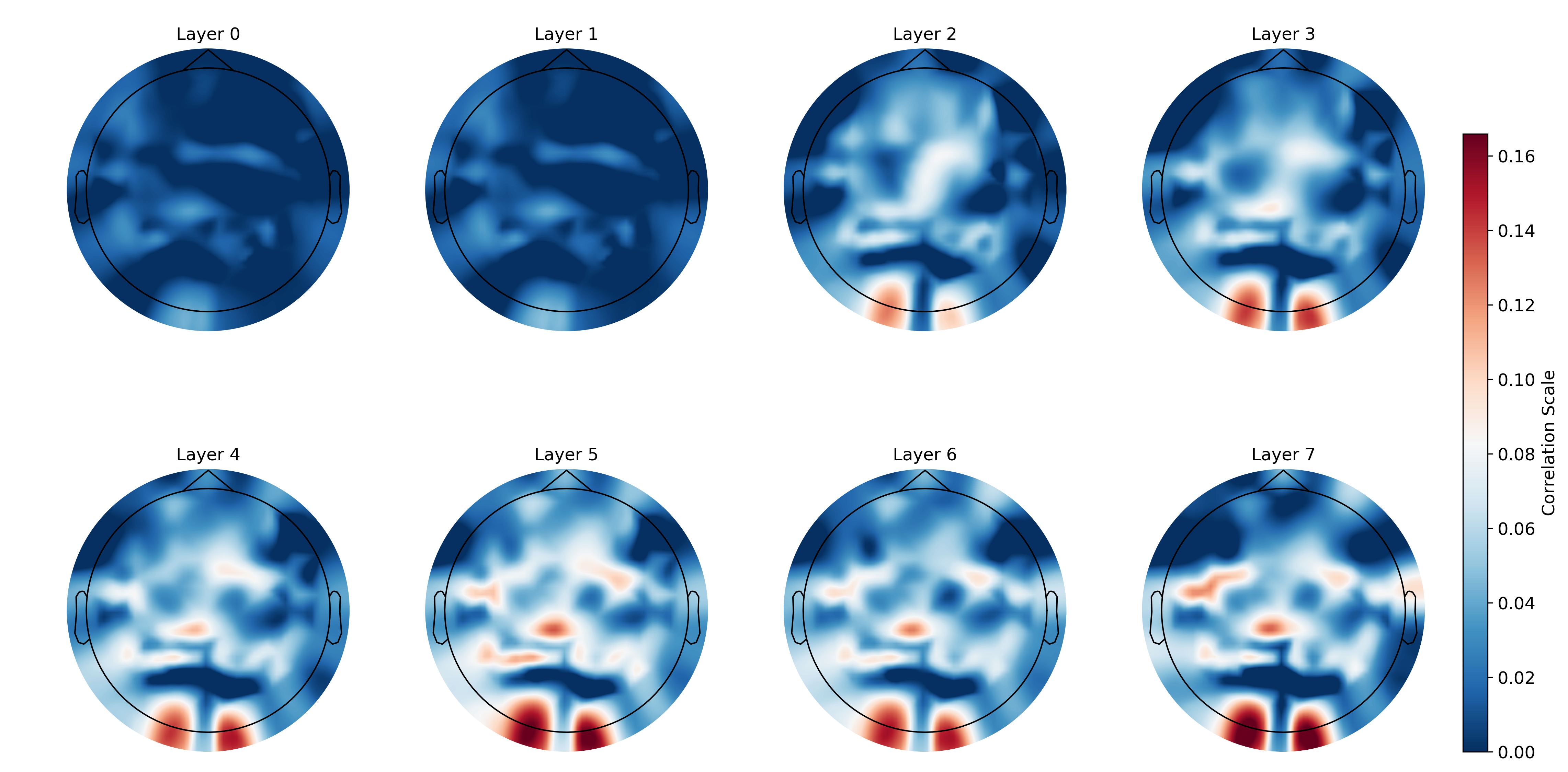}
\caption{Layer-wise ROI S What stream-EEG RSA analysis at peak timepoint. The later layers show ventral-biased occipital-temporal topographies, consistent with object and appearance-focused ventral processing.}\label{figs3}
\end{figure}

\begin{figure}
\centering
\includegraphics[width=0.9\textwidth]{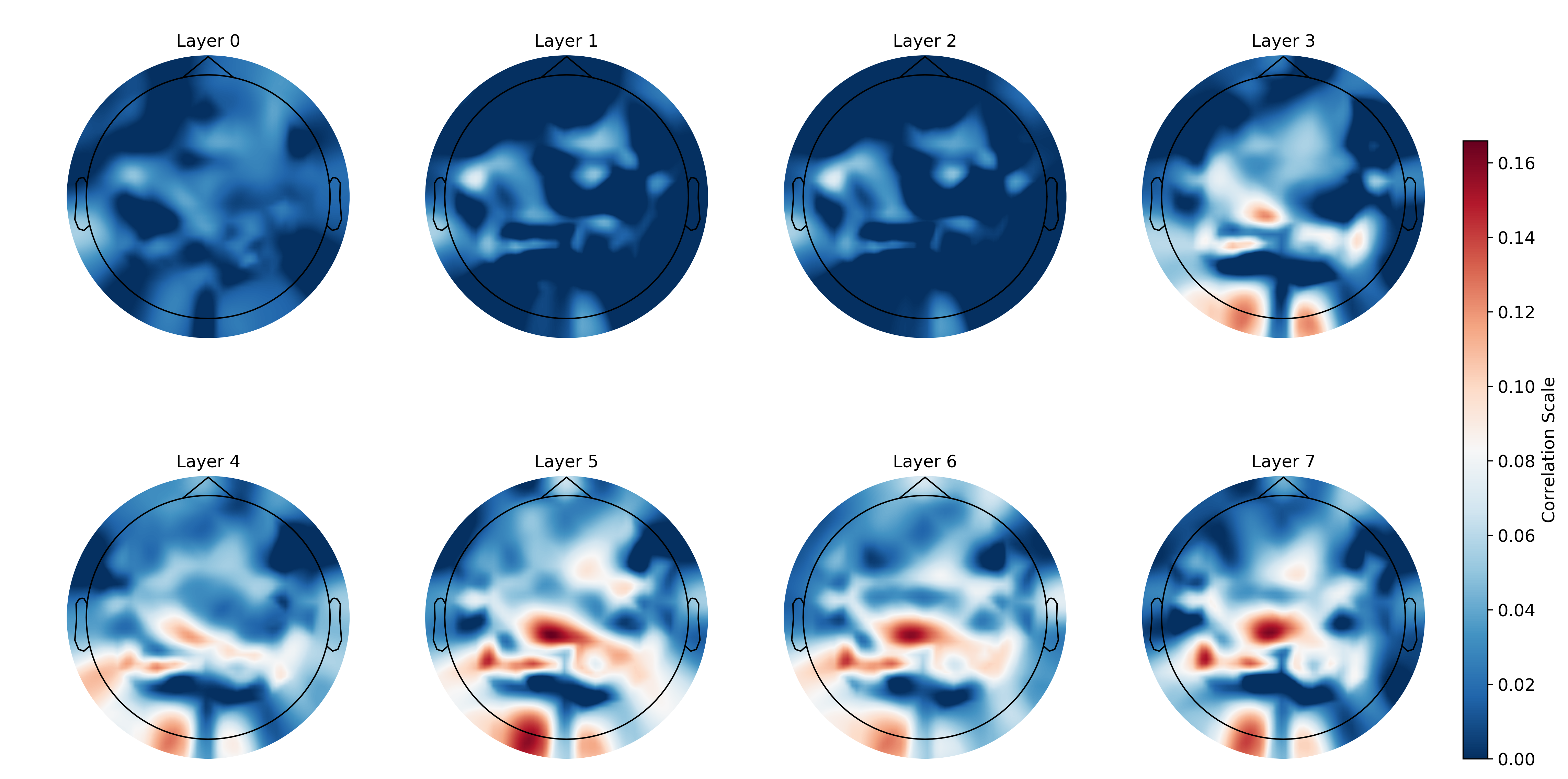}
\caption{Layer-wise ROI S Where stream-EEG RSA analysis at peak timepoint. Later layers show increasingly focal parietal-occipital topographies, highlighting motion and spatially tuned dorsal-like representations.}\label{figs4}
\end{figure}

The scalp topographies at the peak latency, illustrated in Fig.~\ref{figs3} and Fig.~\ref{figs4}, visualize how this alignment is distributed across the head for each layer, revealing a qualitative shift from diffuse, low-amplitude correspondence in early layers to more focal, high-amplitude patterns in occipital, temporal, and parietal regions in deeper layers. In the S Where stream, the strongest correlations emerge over parietal and lateral occipital sensors in higher layers, echoing the putative dorsal pathway’s emphasis on spatial relations and motion processing. In the S What stream, late layers show enhanced correlations over ventral occipital and inferior temporal electrodes, in line with ventral visual areas involved in object recognition. For this paper, these figures therefore provide a layer-resolved bridge between the architectural design (separate what and where streams with selection blocks) and the neurobiological motivation, demonstrating that both streams converge on brain-like representations at behaviorally relevant latencies, each with distinct but complementary spatial signatures that mirror ventral and dorsal pathway organization~\citep{goodale1992separate,creem2001}.

\section{Appendix: EEG Experiment Details}\label{appendixb}

\subsection{Video classes}

We selected 21 action sub-classes from the HVU dataset~\citep{Diba2020} to select the videos used in our EEG experiments: 'riding-or-walking-with-horse', 'cutting-the-grass', 'swimming', 'driving-car', 'playing-violin', 'boating', 'gymnastics', 'eating', 'motorcycling', 'feeding-birds', 'opening-present', 'playing-flute', 'climbing-tree', 'drinking', 'crawling-baby', 'laughing', 'playing-drums', 'cooking', 'petting-cat', 'watering-plants', 'rollerblading'

Although each video may contain additional labels (HVU is multi-class and multi-label), we deliberately chose these actions to span a broad semantic space: they include activities involving animate versus inanimate objects and examples of both high-motion and low-motion scenes. We verified that these 21 classes are semantically diverse, using tools like word-embedding similarity measures, so as to cover a wide range of action concepts in our tasks.

\subsection{Consent form}

To document the ethical safeguards of our EEG experiment, we include a translated version of the participant informed consent form used in the study. This form outlines the purpose of the research, the non-invasive EEG procedures, potential risks and benefits, data confidentiality, compensation, and participants' rights (including voluntary participation and the option to withdraw at any time), and is provided here to facilitate transparent ethical review.

\textbf{1. Purpose of the Study}

I understand that the aim of this study is to compare brain activity in response to static images and dynamic videos using EEG signals. The study seeks to investigate how visual representations in the brain change when transitioning from spatial-only features (static images) to combined spatiotemporal features (videos). By addressing this question, we hope to gain a better understanding of how the brain processes video content, which may inform the development of more accurate and efficient video analysis models. A total of 30 participants will take part in this study, and each individual's contribution is considered valuable in advancing our scientific understanding of brain function.

\textbf{2. Voluntary Participation}

I understand that my participation in this research is entirely voluntary. I am not obligated to participate and will not be denied any medical or diagnostic care if I choose not to take part. My relationship with healthcare providers will not be affected by my decision.

\textbf{3. Right to Withdraw}

I understand that I can withdraw from the study at any time, even after initially agreeing to participate, by informing the researcher. Withdrawing will not affect my access to standard medical services.

\textbf{4. Participation Procedure}

I understand that this study involves EEG recording, a non-invasive and safe neuroimaging method. My participation involves a single session lasting approximately 2 hours during which my brain activity will be recorded while viewing stimuli. The session includes several segments, with breaks and refreshments provided in between. An EEG cap with 128 gel-based electrodes will be placed on my head. I will be asked to attentively view a set of video clips and their corresponding static image frames. Reasonable compensation for time and travel expenses will be provided.

\textbf{5. Potential Benefits}

I acknowledge that while there may be no direct personal benefits, a small financial incentive is offered in appreciation of my participation.

\textbf{6. Risks and Discomforts}

I understand that there are no expected risks or side effects associated with EEG recording. This is a completely safe and painless procedure that involves no magnetic fields or harmful exposure. EEG is widely recognized as a safe method for recording brain activity.

\textbf{7. Confidentiality}

I understand that all personal data will be kept strictly confidential. Only group-level results will be published, and no personally identifiable information will be disclosed.

\textbf{8. Ethics Oversight}

I understand that the research ethics committee may access my data to ensure my rights are being respected and that ethical standards are being followed.

\textbf{9. Costs and Compensation}

I understand that I will not incur any costs related to my participation in this research. Compensation for my time and travel expenses will be provided.

\textbf{10. Contact Information}

I was introduced to the [Name] as the point of contact for any questions or concerns I may have regarding the study. His contact information, as well as that of the principal investigator [Name], has been provided to me.

\textbf{11. In Case of Harm}

I understand that if I experience any physical or psychological issues as a result of participating in this study, the research team will be responsible for covering the costs of treatment and compensation.

\textbf{12. Filing Complaints}

I understand that if I have any concerns or complaints about the study or its staff, I can contact the Researches Ethics Committee.

\textbf{13. Form Copies}

I understand that this consent form has been prepared in two copies, one for myself and one for the research team, both of which are to be signed.

\subsection{Task instructions}

To clarify the behavioral paradigm, the exact task instructions shown to participants on the screen during the EEG recording are reproduced below in translated form. These instructions guided participants through the video and image blocks, explained how to respond to fixation color changes with the space key, and specified when to start, pause, or resume each part of the experiment. 

\textbf{- Introduction}

Hello, and thank you very much for taking the time to participate in this experiment. Press the \textbf{SPACE} key to continue.

\textbf{- Video Task Instructions}

In this section, a series of videos will be presented. Some of the videos are intact, while in others, the order of frames has been disrupted. Additionally, the color of the \textbf{+} symbol at the center of the screen will occasionally change. Please press the \textbf{SPACE} key whenever you notice a color change. Pressing the \textbf{SPACE} key will start the video presentation.

\textbf{- Break}

Thank you for staying with us so far. You may now take a short break before continuing with the task. Press the \textbf{SPACE} key to resume the video presentation.

\textbf{- Image Task Instructions}

In this section, a series of still images will be presented. Please look at each image carefully. Press the \textbf{SPACE} key to start the image presentation.

\textbf{- End}

We sincerely thank you for your participation.

\bibliographystyle{cas-model2-names}

\bibliography{sn-bibliography}

@inproceedings{Bertasius2021,
   author = "Gedas Bertasius and Heng Wang and Lorenzo Torresani",
   issue = "3",
   booktitle = "International Conference on Machine Learning",
   pages = "4",
   title = "Is space-time attention all you need for video understanding?",
   volume = "2",
   year = "2021",
}

@inproceedings{yan2022,
  title= "Multiview transformers for video recognition",
  author= "Yan, Shen and Xiong, Xuehan and Arnab, Anurag and Lu, Zhichao and Zhang, Mi and Sun, Chen and Schmid, Cordelia",
  booktitle= "Proceedings of the IEEE/CVF conference on computer vision and pattern recognition",
  pages= "3333--3343",
  year= "2022"
}

@inproceedings{wang2023,
  title={Videomae v2: Scaling video masked autoencoders with dual masking},
  author={Wang, Limin and Huang, Bingkun and Zhao, Zhiyu and Tong, Zhan and He, Yinan and Wang, Yi and Wang, Yali and Qiao, Yu},
  booktitle={Proceedings of the IEEE/CVF conference on computer vision and pattern recognition},
  pages={14549--14560},
  year={2023}
}

@inproceedings{Liu2022,
   author = {Ze Liu and Jia Ning and Yue Cao and Yixuan Wei and Zheng Zhang and Stephen Lin and Han Hu},
   booktitle = {Proceedings of the IEEE/CVF conference on computer vision and pattern recognition},
   pages = {3202-3211},
   title = {Video swin transformer},
   year = {2022},
}

@inproceedings{Li2022,
   author = {Yanghao Li and Chao-Yuan Wu and Haoqi Fan and Karttikeya Mangalam and Bo Xiong and Jitendra Malik and Christoph Feichtenhofer},
   booktitle = {Proceedings of the IEEE/CVF conference on computer vision and pattern recognition},
   pages = {4804-4814},
   title = {Mvitv2: Improved multiscale vision transformers for classification and detection},
   year = {2022},
}

@inproceedings{liu2022swin,
  title={Swin transformer v2: Scaling up capacity and resolution},
  author={Liu, Ze and Hu, Han and Lin, Yutong and Yao, Zhuliang and Xie, Zhenda and Wei, Yixuan and Ning, Jia and Cao, Yue and Zhang, Zheng and Dong, Li and others},
  booktitle={Proceedings of the IEEE/CVF conference on computer vision and pattern recognition},
  pages={12009--12019},
  year={2022}
}

@inproceedings{wang2024internvideo2,
  title={Internvideo2: Scaling foundation models for multimodal video understanding},
  author={Wang, Yi and Li, Kunchang and Li, Xinhao and Yu, Jiashuo and He, Yinan and Chen, Guo and Pei, Baoqi and Zheng, Rongkun and Wang, Zun and Shi, Yansong and others},
  booktitle={European Conference on Computer Vision},
  pages={396--416},
  year={2024},
  organization={Springer}
}

@inproceedings{zhao2024videoprism,
  title={VideoPrism: A Foundational Visual Encoder for Video Understanding},
  author={Zhao, Long and Gundavarapu, Nitesh Bharadwaj and Yuan, Liangzhe and Zhou, Hao and Yan, Shen and Sun, Jennifer J and Friedman, Luke and Qian, Rui and Weyand, Tobias and Zhao, Yue and others},
  booktitle={International Conference on Machine Learning},
  pages={60785--60811},
  year={2024},
  organization={PMLR}
}

@article{lu2024enhancing,
  title={Enhancing video transformers for action understanding with vlm-aided training},
  author={Lu, Hui and Jian, Hu and Poppe, Ronald and Salah, Albert Ali},
  journal={arXiv preprint arXiv:2403.16128},
  year={2024}
}

@article{assran2025v,
  title={V-jepa 2: Self-supervised video models enable understanding, prediction and planning},
  author={Assran, Mido and Bardes, Adrien and Fan, David and Garrido, Quentin and Howes, Russell and Muckley, Matthew and Rizvi, Ammar and Roberts, Claire and Sinha, Koustuv and Zholus, Artem and others},
  journal={arXiv preprint arXiv:2506.09985},
  year={2025}
}

@inproceedings{tran2019video,
  title={Video classification with channel-separated convolutional networks},
  author={Tran, Du and Wang, Heng and Torresani, Lorenzo and Feiszli, Matt},
  booktitle={Proceedings of the IEEE/CVF international conference on computer vision},
  pages={5552--5561},
  year={2019}
}

@inproceedings{Feichtenhofer2019,
   author = {Christoph Feichtenhofer and Haoqi Fan and Jitendra Malik and Kaiming He},
   booktitle = {Proceedings of the IEEE/CVF international conference on computer vision},
   pages = {6202-6211},
   title = {Slowfast networks for video recognition},
   year = {2019},
}

@inproceedings{Carreira2017,
   author = {Joao Carreira and Andrew Zisserman},
   booktitle = {proceedings of the IEEE Conference on Computer Vision and Pattern Recognition},
   pages = {6299-6308},
   title = {Quo vadis, action recognition? a new model and the kinetics dataset},
   year = {2017},
}

@inproceedings{goyal2017something,
  title={The" something something" video database for learning and evaluating visual common sense},
  author={Goyal, Raghav and Ebrahimi Kahou, Samira and Michalski, Vincent and Materzynska, Joanna and Westphal, Susanne and Kim, Heuna and Haenel, Valentin and Fruend, Ingo and Yianilos, Peter and Mueller-Freitag, Moritz and others},
  booktitle={Proceedings of the IEEE international conference on computer vision},
  pages={5842--5850},
  year={2017}
}

@inproceedings{deng2009imagenet,
  title={Imagenet: A large-scale hierarchical image database},
  author={Deng, Jia and Dong, Wei and Socher, Richard and Li, Li-Jia and Li, Kai and Fei-Fei, Li},
  booktitle={2009 IEEE conference on computer vision and pattern recognition},
  pages={248--255},
  year={2009},
  organization={Ieee}
}

@article{goodale1992separate,
  title={Separate visual pathways for perception and action},
  author={Goodale, Melvyn A and Milner, A David},
  journal={Trends in neurosciences},
  volume={15},
  number={1},
  pages={20--25},
  year={1992},
  publisher={Elsevier}
}

@inproceedings{Arnab2021,
   author = {Anurag Arnab and Mostafa Dehghani and Georg Heigold and Chen Sun and Mario Lučić and Cordelia Schmid},
   booktitle = {Proceedings of the IEEE/CVF international conference on computer vision},
   pages = {6836-6846},
   title = {Vivit: A video vision transformer},
   year = {2021},
}

@article{itti2001computational,
  title={Computational modelling of visual attention},
  author={Itti, Laurent and Koch, Christof},
  journal={Nature reviews neuroscience},
  volume={2},
  number={3},
  pages={194--203},
  year={2001},
  publisher={Nature Publishing Group UK London}
}

@article{desimone1995neural,
  title={Neural mechanisms of selective visual attention},
  author={Desimone, Robert and Duncan, John and others},
  journal={Annual review of neuroscience},
  volume={18},
  number={1},
  pages={193--222},
  year={1995}
}

@article{zhou2000coding,
  title={Coding of border ownership in monkey visual cortex},
  author={Zhou, Hong and Friedman, Howard S and Von Der Heydt, R{\"u}diger},
  journal={Journal of Neuroscience},
  volume={20},
  number={17},
  pages={6594--6611},
  year={2000},
  publisher={Society for Neuroscience}
}

@inproceedings{lin2019tsm,
  title={Tsm: Temporal shift module for efficient video understanding},
  author={Lin, Ji and Gan, Chuang and Han, Song},
  booktitle={Proceedings of the IEEE/CVF international conference on computer vision},
  pages={7083--7093},
  year={2019}
}

@inproceedings{fan2020rubiksnet,
  title={Rubiksnet: Learnable 3d-shift for efficient video action recognition},
  author={Fan, Linxi and Buch, Shyamal and Wang, Guanzhi and Cao, Ryan and Zhu, Yuke and Niebles, Juan Carlos and Fei-Fei, Li},
  booktitle={European Conference on Computer Vision},
  pages={505--521},
  year={2020},
  organization={Springer}
}

@inproceedings{jang2017categorical,
  title={Categorical Reparameterization with Gumbel-Softmax},
  author={Jang, Eric and Gu, Shixiang and Poole, Ben},
  booktitle={International Conference on Learning Representations},
  year={2017}
}

@inproceedings{herrmann2020channel,
  title={Channel selection using gumbel softmax},
  author={Herrmann, Charles and Bowen, Richard Strong and Zabih, Ramin},
  booktitle={European conference on computer vision},
  pages={241--257},
  year={2020},
  organization={Springer}
}

@article{kriegeskorte2008,
  title={Representational similarity analysis-connecting the branches of systems neuroscience},
  author={Kriegeskorte, Nikolaus and Mur, Marieke and Bandettini, Peter A},
  journal={Frontiers in systems neuroscience},
  volume={2},
  pages={249},
  year={2008},
  publisher={Frontiers}
}

@inproceedings{Diba2020,
   author = {Ali Diba and Mohsen Fayyaz and Vivek Sharma and Manohar Paluri and Jürgen Gall and Rainer Stiefelhagen and Luc Van Gool},
   booktitle = {Computer Vision–ECCV 2020: 16th European Conference, Glasgow, UK, August 23–28, 2020, Proceedings, Part V 16},
   pages = {593-610},
   title = {Large scale holistic video understanding},
   year = {2020},
}

@article{Simonyan2014,
   author = {Karen Simonyan and Andrew Zisserman},
   journal = {Advances in neural information processing systems},
   title = {Two-stream convolutional networks for action recognition in videos},
   volume = {27},
   year = {2014},
}

@article{Zareh2024,
   author = {Masoumeh Zareh and Elaheh Toulabinejad and Mohammad Hossein Manshaei and Sayed Jalal Zahabi},
   issue = {1},
   journal = {Scientific Reports},
   pages = {27464},
   publisher = {Nature Publishing Group UK London},
   title = {A deep learning model of dorsal and ventral visual streams for DVSD},
   volume = {14},
   year = {2024},
}

@article{Cichy2014,
   author = {Radoslaw Martin Cichy and Dimitrios Pantazis and Aude Oliva},
   issue = {3},
   journal = {Nature neuroscience},
   pages = {455-462},
   publisher = {Nature Publishing Group US New York},
   title = {Resolving human object recognition in space and time},
   volume = {17},
   year = {2014},
}

@article{fadaei2024,
  title={Going beyond still images to improve input variance resilience in multi-stream vision understanding models},
  author={Fadaei, Amir Hosein and Dehaqani, Mohammad-Reza A},
  journal={Scientific Reports},
  volume={14},
  number={1},
  pages={15366},
  year={2024},
  publisher={Nature Publishing Group UK London}
}

@article{creem2001,
  title={Defining the cortical visual systems:“what”,“where”, and “how”},
  author={Creem, Sarah H and Proffitt, Dennis R},
  journal={Acta psychologica},
  volume={107},
  number={1-3},
  pages={43--68},
  year={2001},
  publisher={Elsevier}
}

@article{yamins2016,
  title={Using goal-driven deep learning models to understand sensory cortex},
  author={Yamins, Daniel LK and DiCarlo, James J},
  journal={Nature neuroscience},
  volume={19},
  number={3},
  pages={356--365},
  year={2016},
  publisher={Nature Publishing Group}
}

@article{schirrmeister2017,
  title={Deep learning with convolutional neural networks for EEG decoding and visualization},
  author={Schirrmeister, Robin Tibor and Springenberg, Jost Tobias and Fiederer, Lukas Dominique Josef and Glasstetter, Martin and Eggensperger, Katharina and Tangermann, Michael and Hutter, Frank and Burgard, Wolfram and Ball, Tonio},
  journal={Human brain mapping},
  volume={38},
  number={11},
  pages={5391--5420},
  year={2017},
  publisher={Wiley Online Library}
}

@article{schrimpf2018,
  title={Brain-score: Which artificial neural network for object recognition is most brain-like?},
  author={Schrimpf, Martin and Kubilius, Jonas and Hong, Ha and Majaj, Najib J and Rajalingham, Rishi and Issa, Elias B and Kar, Kohitij and Bashivan, Pouya and Prescott-Roy, Jonathan and Geiger, Franziska and others},
  journal={BioRxiv},
  pages={407007},
  year={2018},
  publisher={Cold Spring Harbor Laboratory}
}

@article{celeghin2023,
  title={Convolutional neural networks for vision neuroscience: significance, developments, and outstanding issues},
  author={Celeghin, Alessia and Borriero, Alessio and Orsenigo, Davide and Diano, Matteo and M{\'e}ndez Guerrero, Carlos Andr{\'e}s and Perotti, Alan and Petri, Giovanni and Tamietto, Marco},
  journal={Frontiers in Computational Neuroscience},
  volume={17},
  pages={1153572},
  year={2023},
  publisher={Frontiers Media SA}
}

@article{conwell2024large,
  title={A large-scale examination of inductive biases shaping high-level visual representation in brains and machines},
  author={Conwell, Colin and Prince, Jacob S and Kay, Kendrick N and Alvarez, George A and Konkle, Talia},
  journal={Nature communications},
  volume={15},
  number={1},
  pages={9383},
  year={2024},
  publisher={Nature Publishing Group UK London}
}

@article{soni2024conclusions,
  title={Conclusions about neural network to brain alignment are profoundly impacted by the similarity measure},
  author={Soni, Ansh and Srivastava, Sudhanshu and Kording, Konrad and Khosla, Meenakshi},
  journal={bioRxiv},
  pages={2024--08},
  year={2024},
  publisher={Cold Spring Harbor Laboratory}
}

@article{Rao2021,
   author = {Yongming Rao and Wenliang Zhao and Benlin Liu and Jiwen Lu and Jie Zhou and Cho-Jui Hsieh},
   journal = {Advances in neural information processing systems},
   pages = {13937-13949},
   title = {Dynamicvit: Efficient vision transformers with dynamic token sparsification},
   volume = {34},
   year = {2021},
}

@inproceedings{Liu2021Swin,
   author = {Ze Liu and Yutong Lin and Yue Cao and Han Hu and Yixuan Wei and Zheng Zhang and Stephen Lin and Baining Guo},
   booktitle = {Proceedings of the IEEE/CVF international conference on computer vision},
   pages = {10012-10022},
   title = {Swin transformer: Hierarchical vision transformer using shifted windows},
   year = {2021},
}

@inproceedings{fan2021Multiscale,
  title={Multiscale vision transformers},
  author={Fan, Haoqi and Xiong, Bo and Mangalam, Karttikeya and Li, Yanghao and Yan, Zhicheng and Malik, Jitendra and Feichtenhofer, Christoph},
  booktitle={Proceedings of the IEEE/CVF international conference on computer vision},
  pages={6824--6835},
  year={2021}
}

@inproceedings{shazeer2017,
  title={Outrageously Large Neural Networks: The Sparsely-Gated Mixture-of-Experts Layer},
  author={Shazeer, Noam and Mirhoseini, Azalia and Maziarz, Krzysztof and Davis, Andy and Le, Quoc and Hinton, Geoffrey and Dean, Jeff},
  booktitle={International Conference on Learning Representations},
  year={2017}
}

@article{beltagy2020,
  title={Longformer: The long-document transformer},
  author={Beltagy, Iz and Peters, Matthew E and Cohan, Arman},
  journal={arXiv preprint arXiv:2004.05150},
  year={2020}
}

@inproceedings{kitaev2020,
  title={Reformer: The Efficient Transformer},
  author={Kitaev, Nikita and Kaiser, Lukasz and Levskaya, Anselm},
  booktitle={International Conference on Learning Representations},
  year={2020}
}

@inproceedings{tschannen2020self,
  title={Self-supervised learning of video-induced visual invariances},
  author={Tschannen, Michael and Djolonga, Josip and Ritter, Marvin and Mahendran, Aravindh and Houlsby, Neil and Gelly, Sylvain and Lucic, Mario},
  booktitle={Proceedings of the IEEE/CVF Conference on Computer Vision and Pattern Recognition},
  pages={13806--13815},
  year={2020}
}

@inproceedings{li2023simple,
  title={A simple baseline for video restoration with grouped spatial-temporal shift},
  author={Li, Dasong and Shi, Xiaoyu and Zhang, Yi and Cheung, Ka Chun and See, Simon and Wang, Xiaogang and Qin, Hongwei and Li, Hongsheng},
  booktitle={Proceedings of the IEEE/CVF Conference on Computer Vision and Pattern Recognition},
  pages={9822--9832},
  year={2023}
}

@inproceedings{zhang2021token,
  title={Token shift transformer for video classification},
  author={Zhang, Hao and Hao, Yanbin and Ngo, Chong-Wah},
  booktitle={Proceedings of the 29th acm international conference on multimedia},
  pages={917--925},
  year={2021}
}

@inproceedings{dapello2022,
  title={Aligning model and macaque inferior temporal cortex representations improves model-to-human behavioral alignment and adversarial robustness.. bioRxiv 2022.07. 01.498495},
  author={Dapello, J and Kar, K and Schrimpf, M and Geary, RB and Ferguson, M and Cox, DD and DiCarlo, JJ},
  booktitle={International Conference on Learning Representations},
  year={2023}
}

@article{fisher1936,
  title={The use of multiple measurements in taxonomic problems},
  author={Fisher, Ronald A},
  journal={Annals of eugenics},
  volume={7},
  number={2},
  pages={179--188},
  year={1936},
  publisher={Wiley Online Library}
}

@article{vaswani2017attention,
  title={Attention is all you need},
  author={Vaswani, Ashish and Shazeer, Noam and Parmar, Niki and Uszkoreit, Jakob and Jones, Llion and Gomez, Aidan N and Kaiser, {\L}ukasz and Polosukhin, Illia},
  journal={Advances in neural information processing systems},
  volume={30},
  year={2017}
}

@article{dosovitskiy2020image,
  title={An Image is Worth 16x16 Words: Transformers for Image Recognition at Scale},
  author={Dosovitskiy, Alexey and Beyer, Lucas and Kolesnikov, Alexander and Weissenborn, Dirk and Zhai, Xiaohua and Unterthiner, Thomas and Dehghani, Mostafa and Minderer, Matthias and Heigold, Georg and Gelly, Sylvain and others},
  booktitle={International Conference on Learning Representations}
}

@article{wang2020linformer,
  title={Linformer: Self-attention with linear complexity},
  author={Wang, Sinong and Li, Belinda Z and Khabsa, Madian and Fang, Han and Ma, Hao},
  journal={arXiv preprint arXiv:2006.04768},
  year={2020}
}

@inproceedings{choromanski2021rethinking,
  title={Rethinking Attention with Performers},
  author={Choromanski, Krzysztof Marcin and Likhosherstov, Valerii and Dohan, David and Song, Xingyou and Gane, Andreea and Sarlos, Tamas and Hawkins, Peter and Davis, Jared Quincy and Mohiuddin, Afroz and Kaiser, Lukasz and others},
  booktitle={International Conference on Learning Representations}
}

@article{zaheer2020big,
  title={Big bird: Transformers for longer sequences},
  author={Zaheer, Manzil and Guruganesh, Guru and Dubey, Kumar Avinava and Ainslie, Joshua and Alberti, Chris and Ontanon, Santiago and Pham, Philip and Ravula, Anirudh and Wang, Qifan and Yang, Li and others},
  journal={Advances in neural information processing systems},
  volume={33},
  pages={17283--17297},
  year={2020}
}

@article{child2019generating,
  title={Generating long sequences with sparse transformers},
  author={Child, Rewon and Gray, Scott and Radford, Alec and Sutskever, Ilya},
  journal={arXiv preprint arXiv:1904.10509},
  year={2019}
}

@article{ryoo2021tokenlearner,
  title={Tokenlearner: What can 8 learned tokens do for images and videos?},
  author={Ryoo, Michael S and Piergiovanni, AJ and Arnab, Anurag and Dehghani, Mostafa and Angelova, Anelia},
  journal={arXiv preprint arXiv:2106.11297},
  year={2021}
}

@article{cichy2019deep,
  title={Deep neural networks as scientific models},
  author={Cichy, Radoslaw M and Kaiser, Daniel},
  journal={Trends in cognitive sciences},
  volume={23},
  number={4},
  pages={305--317},
  year={2019},
  publisher={Elsevier}
}

@article{kietzmann2019recurrence,
  title={Recurrence is required to capture the representational dynamics of the human visual system},
  author={Kietzmann, Tim C and Spoerer, Courtney J and S{\"o}rensen, Lynn KA and Cichy, Radoslaw M and Hauk, Olaf and Kriegeskorte, Nikolaus},
  journal={Proceedings of the National Academy of Sciences},
  volume={116},
  number={43},
  pages={21854--21863},
  year={2019},
  publisher={National Academy of Sciences}
}

@article{lawhern2018eegnet,
  title={EEGNet: a compact convolutional neural network for EEG-based brain--computer interfaces},
  author={Lawhern, Vernon J and Solon, Amelia J and Waytowich, Nicholas R and Gordon, Stephen M and Hung, Chou P and Lance, Brent J},
  journal={Journal of neural engineering},
  volume={15},
  number={5},
  pages={056013},
  year={2018},
  publisher={iOP Publishing}
}



\end{document}